\documentclass[final]{cvpr}

\usepackage{times}
\usepackage{epsfig}
\usepackage{graphicx}
\usepackage{amsmath}
\usepackage{amssymb}
\usepackage{bm}
\usepackage{multirow}
\usepackage{cuted}
\usepackage{capt-of}
\usepackage{xcolor}
\usepackage{enumitem}
\usepackage[pagebackref=true,breaklinks=true,letterpaper=true,colorlinks,bookmarks=false]{hyperref}

\DeclareMathOperator*{\argmin}{arg\,min}

\begin{document}

\graphicspath{{images/}}

\newcommand{\bx}{{\bf x}}
\newcommand{\bz}{{\bf z}}
\newcommand{\balpha}{{\bm \alpha}}
\newcommand{\bbeta}{{\bm \beta}}
\newcommand{\bdelta}{{\bm \delta}}
\newcommand{\bgamma}{{\bm \gamma}}
\newcommand{\btheta}{{\bm \theta}}
\newcommand{\blambda}{{\bm \lambda}}
\newcommand{\bS}{{\bf S}}
\newcommand{\bT}{{\bf T}}
\newcommand{\bB}{{\bf B}}
\newcommand{\tmp}{u}

\newcommand{\paravspace}{\vspace{-10pt}}
\newcommand{\eqvspace}{\vspace{-3pt}}

\vspace{-20pt}
\title{\!\!\!{\spaceskip=0.1855em\relax Deformed Implicit Field: Modeling 3D Shapes with Learned Dense Correspondence}\!\!\!}

\author{Yu Deng\thanks{This work was done when Yu Deng was an intern at MSRA.}\,\,$^{1,2}$ \quad Jiaolong Yang$^{2}$ \quad Xin Tong$^{2}$ \\
	$^1${Tsinghua University} \quad  $^2${Microsoft Research Asia} \\
	{\tt\small \{t-yudeng,jiaoyan,xtong\}@microsoft.com}
}

\maketitle
\begin{strip}
	\vspace{-40pt}
	\centering
	\includegraphics[width=\textwidth]{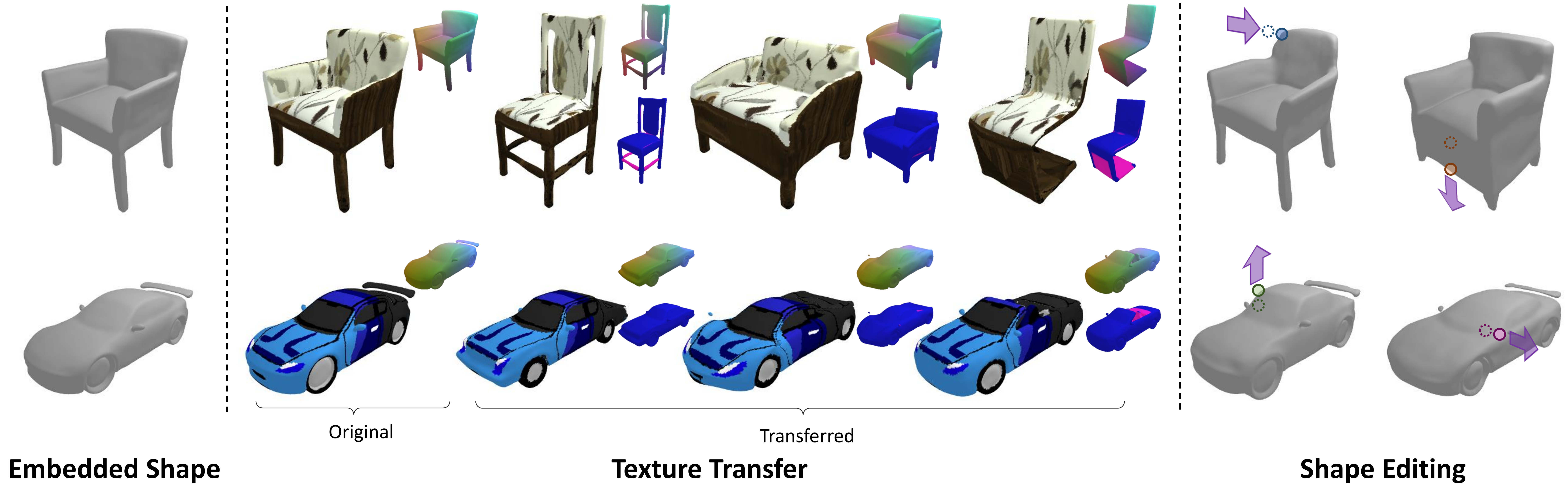}
	\vspace{-10pt}
	\captionof{figure}{Our DIF-Net can produce 3D shapes with dense correspondences for object categories containing complex geometry variation and structure differences. It enables high-quality texture transfer shown in the middle four columns, where the two smaller figures after each transfer result show the color-coded  correspondences (top) and their uncertainty (bottom; blue and red indicates low and high uncertainty respectively). 
	With our learned shape space and correspondence, shapes can be freely edited by simply moving one or a sparse set of points, as shown in the last two columns.
	\label{fig:teaser}}
	\vspace{2.5pt}
\end{strip}
\begin{abstract}
	\vspace{-5pt}
	We propose a novel Deformed Implicit Field (DIF) representation for modeling 3D shapes of a category and generating dense correspondences among shapes. With DIF, a 3D shape is represented by a template implicit field shared across the category, together with a 3D deformation field and a correction field dedicated for each shape instance. Shape correspondences can be easily established using their deformation fields.
	Our neural network, dubbed DIF-Net, jointly learns a shape latent space and these fields for 3D objects belonging to a category without using any correspondence or part label. The learned DIF-Net can also provides reliable correspondence uncertainty measurement reflecting shape structure discrepancy. 
	Experiments show that DIF-Net not only produces high-fidelity 3D shapes but also builds high-quality dense correspondences across different shapes. We also demonstrate several applications such as texture transfer and shape editing, where our method achieves compelling results that cannot be achieved by  previous methods. \footnote{Code URL: \href{https://github.com/microsoft/DIF-Net}{https://github.com/microsoft/DIF-Net}.}
	
\end{abstract}

\vspace{-17pt}
\section{Introduction}
3D objects in a same class share some common shape features and semantic correspondences, which can be used to construct a deformable shape model beneficial for a diverse array of downstream tasks in 3D and 2D domains such as shape understanding~\cite{lassner2017unite,alp2018densepose}, reconstruction~\cite{tewari2017mofa,cao2015real,zuffi2018lions}, manipulation~\cite{blanz1999morphable,han2017deepsketch2face}, and image synthesis~\cite{thies2016face2face,weng2019photo,tewari2020stylerig}.

Learning a 3D shape model with dense correspondences is a longstanding task in computer vision and graphics. However, existing works mostly focus on object classes with consistent geometric topologies such as human face and body~\cite{blanz1999morphable,vlasic2006face,li2017learning,loper2015smpl,zuffi20173d}. Shapes in these object categories can be pre-aligned for 3D model construction. Recent deep learning based approaches directly learn a latent space of 3D objects~\cite{wu2016learning,tatarchenko2017octree,achlioptas2018learning,park2019deepsdf}. Although these methods can model complex objects, they do not deal with dense correspondences between 3D shapes. 

In this paper, we investigate learning model of 3D shapes and their dense correspondences for more generic objects such as cars and chairs.
Compared to human face and body, these object classes exhibit much larger shape variations and structure changes, rendering correspondence construction extremely challenging. For these object categories, even human cannot reliably label the dense correspondences between two arbitrary shapes.

To achieve this goal, we act on recent advances in deep implicit fields, which have shown extraordinary power of representing complicated 3D geometry~\cite{mescheder2019occupancy,park2019deepsdf,chen2019learning,sitzmann2020implicit}, and propose a novel Deformed Implicit Field (DIF) representation for joint shape latent space and dense correspondence learning. 
With DIF, a 3D shape is represented by a template implicit field, shared across the category, together with a 3D deformation field and a scalar correction field, dedicated for each shape instance. The output implicit field of a shape can be constructed by deforming the template implicit field and applying correction.
The deformation field serves as a shape alignment function, with which dense correspondence between two shapes can be established by deforming their surfaces to the aligned 3D space. 
The correction field is introduced to handle structure variations among shapes.

We apply a neural network called DIF-Net to learn these fields together given a collection of shapes. To achieve unsupervised correspondence learning without any label, our key observation is that the normal direction of a shape point is highly correlated to its semantic information and very useful for correspondence reasoning.
In light of this, we simply enforce the normals of two corresponding points connected by deformation to be close. In addition, we impose a spatial smoothness constrain on the deformation fields and enforce the correction fields to be minimal to make it only focus on dealing with structure variations.
Thorough qualitative and quantitative evaluations show that our DIF-Net trained in this way can produce high quality correspondences. Moreover, correspondence uncertainty reflecting structure discrepancy between two shapes can be reliably measured by our method.

\textbf{The contribution of this paper} is as follows:
\vspace{-3pt}
\begin{itemize}
\item We propose DIF, a novel implicit field based 3D shape representation for shapes of an object category.
\vspace{-3pt}
\item We propose DIF-Net, the first method devoted to 3D shape modeling with dense correspondences learned in an unsupervised fashion for objects with structure variation. 
\vspace{-3pt}
\item We show that our method can achieve high-quality dense correspondences and compelling texture transfer and shape editing results that cannot be achieved by previous methods.
We believe our method can be applied in a wide range of 3D shape analysis and manipulation tasks.
\end{itemize}

\section{Related Work}
\paragraph{3D Shape Models with Correspondence.}
Building 3D shape model for a class of shapes has been actively studied in the past.
Perhaps the most famous 3D shape model is the 3D morphable model (3DMM) introduced by Blanz \etal~\cite{blanz1999morphable} for human faces. To build a 3DMM model, face scans are aligned by shape registration methods to derive correspondences, based on which shape deformation bases can be obtained via PCA. The 3DMM model has brought a profound impact to human face related research~\cite{paysan20093d,zhu2015high,tran2018nonlinear,hu2016face,liu2018disentangling,tewari2017mofa,genova2018unsupervised,garrido2016reconstruction,feng2018joint,thies2016face2face,tewari2020stylerig,deng2020disentangled}.
Apart from face, Loper \etal~\cite{loper2015smpl} build a 3D morphable model for skinned human body which can control body shapes and poses. This model has been applied in various tasks such as pose estimation \cite{bogo2016keep,arnab2019exploiting} and image manipulation \cite{weng2019photo}. Similarly, Zuffi \etal~\cite{zuffi20173d} propose a morphable model for animals. 
The object categories handled by these methods typically have consistent topologies where shapes can be aligned to build correspondences. They do not address more complex object classes containing structure variations.

\paravspace
\paragraph{Learning Shape Latent Space.}  A large volume of methods~\cite{wu2016learning,tatarchenko2017octree,achlioptas2018learning,ben2018multi,bagautdinov2018modeling,groueix2018papier,zhu2018visual,tan2018variational,gao2019sdm,park2019deepsdf} have been proposed in recent years to model 3D shapes and learn a latent shape space using deep neural networks, especially generative adversarial networks (GANs) \cite{goodfellow2014generative} and variational auto-encoders \cite{kingma2013auto,tan2018variational}. However these methods do not explicitly model the dense correspondence among different shapes. Our method not only learns a shape latent space but also generates dense shape correspondence.

\paravspace
\paragraph{Implicit Shape Representation.} Recent studies show that learning implicit functions for 3D shapes excels at representing complicated geometry \cite{park2019deepsdf,mescheder2019occupancy,chen2019learning,genova2019learning,genova2020local,atzmon2020sal,atzmon2020sal++,sitzmann2020implicit,duan2020curriculum,hao2020dualsdf}. For example, Park \etal~\cite{park2019deepsdf} use a neural network to approximate the signed distance field (SDF) of 3D shapes and show superior results compare to voxel and mesh based representations~\cite{tatarchenko2017octree,groueix2018papier}. Sitzmann \etal~\cite{sitzmann2020implicit} show that surfaces of complex scene can be represented by a simple 5-layer MLP with periodic functions as activation. However, these methods mainly target at high-fidelity surface reconstruction and cannot reveal shape correspondence. Genova \etal~\cite{genova2019learning} introduce an implicit template constructed with multiple RBF kernels. They can obtain a coarse dense correspondence between shapes by deforming and relocating RBF kernels to fit different shapes. However, their learning process is designed only for shape reconstruction thus the obtained correspondences are not reliable. Our representation in this paper is also based on implicit fields, but enables correspondence reasoning. Our new loss functions leads to high-quality correspondences learned without any label. A concurrent work from Zheng \etal~\cite{zheng2020dit} also uses deformed implicit field for shape modeling. Nevertheless, they do not consider structure differences between shapes.

\paravspace
\paragraph{Structured Shape Representation.} 
Structured representations are also widely used to model complex shapes with varying structures \cite{tulsiani2017learning,deng2020cvxnet,smirnov2020deep,chen2020bsp,mo2019partnet,gao2019sdm,paschalidou2019superquadrics,chen2019bae,hao2020dualsdf}.
By decomposing 3D shapes into small parts, a complicated shape can be represented by primitive elements such as cuboids \cite{tulsiani2017learning,smirnov2020deep}, superquadrics \cite{paschalidou2019superquadrics}, convexes \cite{deng2020cvxnet,chen2020bsp}, and RBF kernels~\cite{genova2019learning}. 
Many of these methods can provide part-level correspondences among shapes, but do not model dense correspondences. Recently, a concurrent work~\cite{liu2020learning} also predicts dense correspondence for 3D shapes with structure variations based on the part-based implicit representation of \cite{chen2019bae}. In contrast, our method does not divide a shape into parts. We obtain dense correspondence via shape deformation, whereas \cite{liu2020learning} achieves this by using a part-aware embedding space. 

\paravspace
\paragraph{Shape Deformation.} Shape deformation~\cite{sederberg1986free,schaefer2006image,joshi2007harmonic,chao2010simple,uy2020deformation,yifan2020neural,jiang2020shapeflow} aims at deforming a shape to best fit a target while preserving local geometric details. Although our method also models shape deformation, our goal is to design a novel shape representation for surface reconstruction as well as correspondence reasoning. Besides, our implicit field based method can handle structure mismatch, which is problematic for previous mesh deformation methods.



\section{Approach}
\subsection{Overview}
Given a collection of 3D objects $\{\mathcal{O}_i\}$ from one category, our goal is to learn a latent shape space $\mathcal{L}$ as well as a neural shape model $f$ that can generate these objects and provide dense shape correspondence. Each shape can be represented by a latent code $\alpha \in \mathbb{R}^k$ in $\mathcal{L}$, and the shape model $f$ maps the latent code to corresponding 3D shape,
\eqvspace
\begin{equation}
f: \alpha \in \mathbb{R}^{k} \rightarrow \mathcal{O}
\eqvspace
\end{equation}
\noindent with a neural network. We adopt the auto-decoder framework presented in \cite{park2019deepsdf} to jointly learn the shape codes $\{\alpha_j\}$ for the given objects and the weights of model $f$. This auto-decoder framework can give rise to a decent latent space as shown in \cite{park2019deepsdf}. 

After training, new shapes can be generated by latent space sampling, and a shape can be embedded into the latent shape via inverse optimization.

\paravspace
\paragraph{Implicit Field.} To generate high-fidelity shapes, we use signed distance fields (SDF) which can faithfully represent surface geometry details using a neural network as the field function \cite{park2019deepsdf}.
SDF is a continuous representation which assigns any point $p\in\mathbb{R}^3$ a scalar value $s\in\mathbb{R}$:
\eqvspace
\begin{equation}
SDF(p) = s,
\eqvspace
\end{equation}
\noindent where the magnitude of $s$ represents the distance from $p$ to its closest shape surface and the sign indicates whether $p$ is inside (negative) the shape or outside (positive). With an SDF, shape surface can be implicitly represented by the iso-surface of $SDF(\cdot)=0$. A 3D mesh can be extracted from this implicit surface using off-the-shelf algorithms such as Marching Cubes~\cite{lorensen1987marching}.
Using SDF to represent shapes, our neural shape model can be rewritten as 
\eqvspace
\begin{equation}
f: (\alpha, p) \in \mathbb{R}^{k+3} \rightarrow s \in \mathbb{R}.\label{eq:shape_model}
\end{equation}

\begin{figure}[t!]
	\small
	\centering
	\includegraphics[width=1.0\columnwidth]{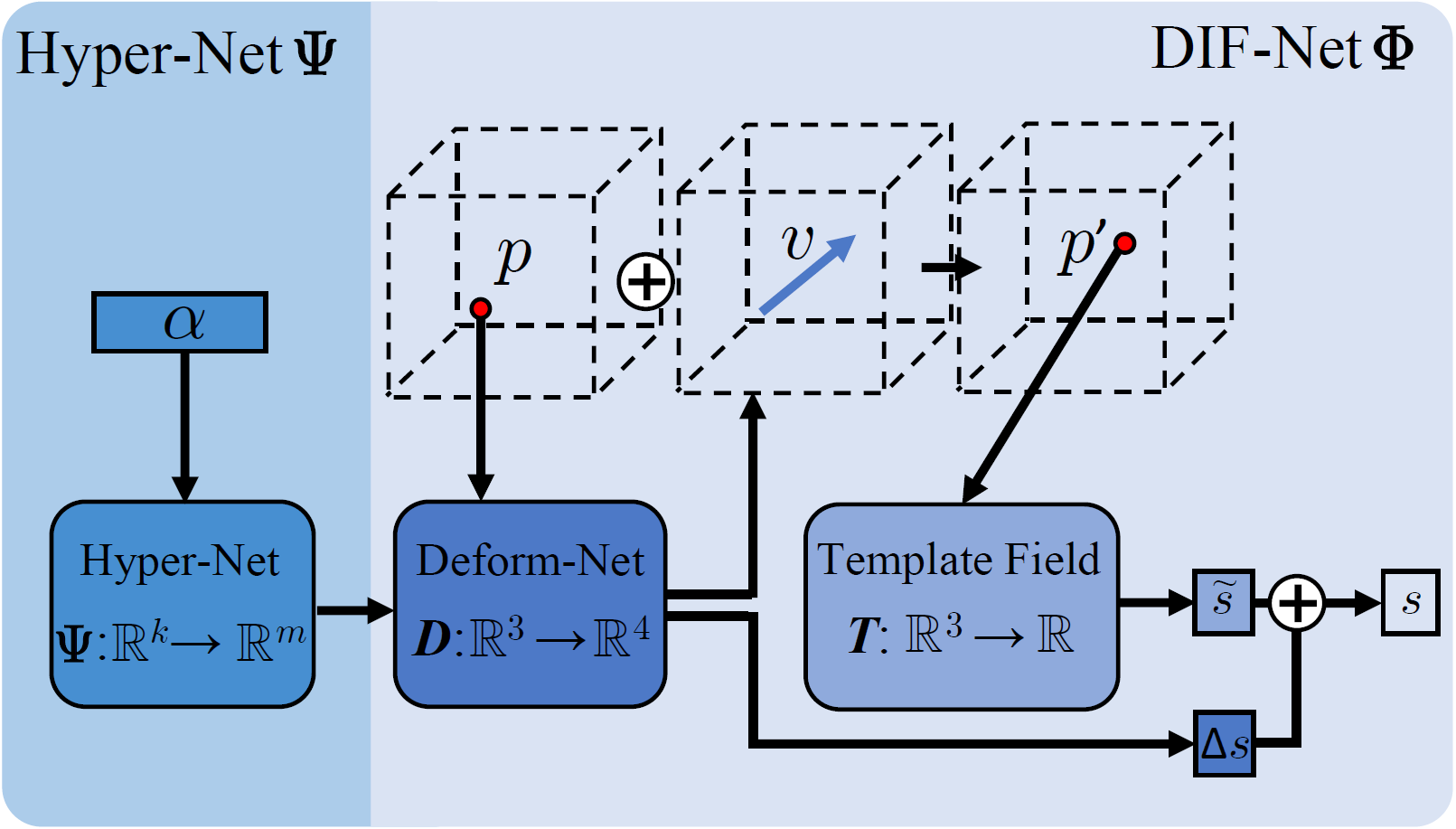}
	\caption{Overview of our proposed method. For a shape code $\alpha$, Hyper-Net $\Psi$ predicts (a part of) the weights of DIF-Net $\Phi$, which further predicts the SDF for the shape. DIF-Net $\Phi$ consists of Deform-Net $D$ which predicts a 3D deformation field and a correction field for the shape, and network $T$ for generating a template implicit field shared across all shapes.}
	\label{fig:framework}
	\vspace{-1pt}
\end{figure}

\paravspace
\paragraph{Network Components.} The task in Eq.~\eqref{eq:shape_model} intertwines shape information decoding from latent codes and SDF prediction for 3D points. Inspired by \cite{sitzmann2019scene}, we employ two networks to decompose this task: a \emph{DIF-Net} for SDF prediction and a \emph{Hyper-Net} for shape information decoding, as illustrated in Fig.~\ref{fig:framework}. Hyper-Net $\Psi$ predicts the weights $\omega$ for the neurons in DIF-Net $\Phi$, and the two networks can be written as
\eqvspace
\begin{equation}
\Psi: \alpha\in\mathbb{R}^k \to \omega\in \mathbb{R}^m,
\end{equation}
\begin{equation}
\Phi_{\omega}: p\in\mathbb{R}^3 \to s\in\mathbb{R}. \label{eq:SDF2}
\eqvspace
\end{equation}
\noindent 
Our DIF-Net $\Phi$ consists of two sub-networks: a template SDF generation network $T$ and a Deform-Net $D$. We will introduce these two sub-networks and our deformed implicit field representation in the next section.

\subsection{Deformed Implicit Field Representation} \label{sec:dif}
For a given object class, we assume that the object instances are mostly composed by a few common patterns or semantic structures. This is a mild assumption valid for many real-world object classes. For example, all cars consist of bodies and tires, and most chairs have back, seat, and legs.
We seek to find a template implicit field which depicts common structures of the class and can derive SDFs for different objects through 3D deformation and correction.

\paravspace
\paragraph{Template Implicit Field.}
To capture common structures of an object category, we learn a template SDF generation network $T$:
\eqvspace
\begin{equation}
T: p\in\mathbb{R}^3 \to \widetilde{s} \in \mathbb{R}, \label{eq:templatesdf-net}
\eqvspace
\end{equation}
which maps a 3D point $p$ to a scalar value $\widetilde{s}$. The latter is used to construct the SDF for a specific object via deformation and correction, which will be described later. The network weights of $T$ is shared across the whole class therefore it is enforced to learn common patterns within the class. 

\begin{figure}[t]
	\small
	\centering
	\includegraphics[width=1.0\columnwidth]{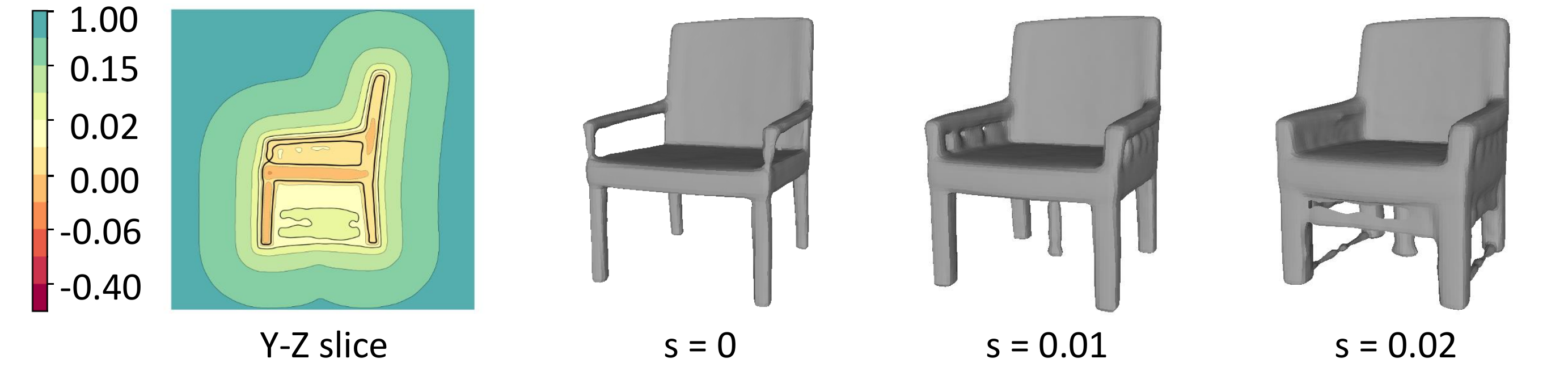}
	\caption{The learned template implicit field (\emph{Y-Z} slice) for chair category and different iso-surfaces extracted from it. Note that the template implicit field is not a valid shape SDF and it characterizes different shape structures within a category. 
	\label{fig:mean}}
	\vspace{-2pt}
\end{figure}

Note that although the SDF of an object can be derived from $T$, $T$ itself need not to be a valid SDF of any certain object. In fact, Figure~\ref{fig:mean} shows that the values of $T$ in the free space are “fused” SDF values of different shapes in the object category but not distances to its iso-surface $s=0$ (\eg, compare the lower part of chair's template field slice and the extracted iso-surface at $s=0$). This makes our template field fundamentally different from previous mesh-based templates~\cite{groueix2018papier,yang2018foldingnet}. Our template field records all structure variations in a category via its different iso-surfaces in the volume while a mesh-based template is only meaningful at its surface. This difference enables our method to learn better correspondence for objects with structure discrepancies, as we will show in the experiments.

\begin{figure}[t!]
	\small
	\centering
	\includegraphics[width=1.0\columnwidth]{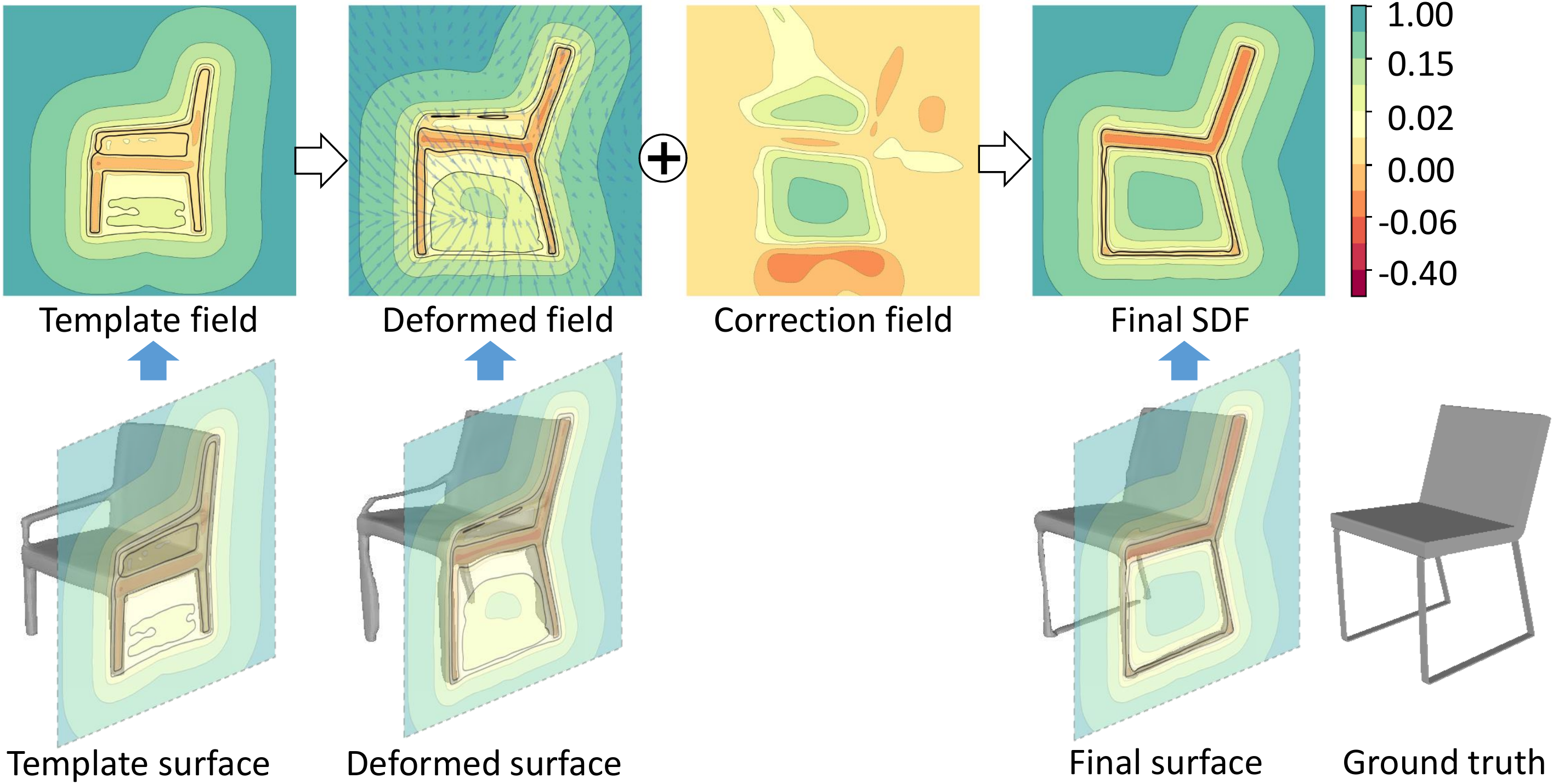}
	\caption{2D and 3D visualization of our SDF prediction process.\label{fig:vis}}
	\vspace{-2pt}
\end{figure}

\paravspace
\paragraph{Deformation Field and Correction Field.} 
To obtain the SDF for a certain object, we learn a Deform-Net $D$ to predict a deformation field as well as a correction field on top of the template field $T$:
\eqvspace
\begin{equation}
D_\omega: p\in\mathbb{R}^3 \to (v, \Delta s) \in \mathbb{R}^4, \label{eq:deform-net}
\eqvspace
\end{equation}
\noindent where $v\in\mathbb{R}^3$ is a deformation flow and $\Delta s\in\mathbb{R}$ is a scalar correction. The weights of $D$ are instance-specific and derived from the Hyper-Net, as shown in Fig.~\ref{fig:framework} and indicated by subscript $\omega$ in Eq.~\eqref{eq:deform-net}.
With $T$ and $D_\omega$, the SDF value of point $p$ of an object can be obtained via
\eqvspace
\begin{equation}
s = T(p + v) + \Delta s = T(p+D^v_\omega(p)) + D^{\Delta s}_\omega(p). \label{eq:sdf_by_deform_correction}
\eqvspace
\end{equation}
The per-point offset $v$ deforms a point $p$ to the template space to obtain its SDF value via the template field $T$, which naturally induces dense correspondence between an object instance and the template field. Dense correspondences between two shapes can be further established by deforming their surface to the aligned template space and then searching nearest neighbors. The per-point correction $\Delta s$ modifies the assigned SDF value of $p$ if it still differs from the ground truth value. The correction field therefore helps to add or delete structures to enhance the shape representation ability, as illustrated in Fig.~\ref{fig:vis}. 

Note that \textbf{\emph{the correction field is also crucial to learn reasonable correspondences under structure variations}}. Consider the chair example in Fig.~\ref{fig:vis}, to generate the side stretcher which does not exist in the template surface using only deformation, some surface points must be deformed to the desired positions, necessitating a complex deformation field that is difficult to learn by network $D_\omega$. 
Even if such a deformation field can be learned, the correspondences derived from it for the stretcher part are wrong. 
In contrast, with a correction field, the stretcher structure can be added by simply altering the SDF values in this region rather than deforming existing surface points. This way, not only wrong correspondences can be avoided, but also a simple, smooth deformation field can be easily learned by the network. The ablation study in Sec.~\ref{sec:ablation} also shows the effectiveness of our correction field.

In summary, our neural shape model can be written as 
\eqvspace
\begin{equation}
f(\alpha, p) \!=\! \Phi_{\Psi(\alpha)}(p) \!=\! T(p\!+\!D^{v}_{\Psi(\alpha)}(p)) \!+\! D^{\Delta s}_{\Psi(\alpha)}(p),\!\!\!\! \label{eq:}
\eqvspace
\end{equation}
\noindent which is parameterized by the weights of network $\Psi$ and $T$.

\subsection{Learning Shape and Correspondence with DIF}

We use the auto-decoder framework presented in \cite{park2019deepsdf} to jointly train weights of networks $\Psi$ and $T$ and learn latent codes $\{\alpha_j\}$. We design new loss functions for DIF to learn desirable dense correspondences. 

Given a collection of shapes, we first apply an SDF regression loss similar to \cite{sitzmann2020implicit} to learn the SDFs of these shapes. Let $\Phi_i(p)$ be the short-hand notation for $\Phi_{\Psi(\alpha_i)}(p)$ which is the predicted SDF value, we have
\begin{equation}\label{loss:sdf}
\begin{split}
\!\!\!\!L_{sdf} &\!=\! \sum_{i}\!\Big(\sum\limits_{p\in\Omega}| \Phi_i(p) - \bar{s}| + \sum\limits_{p\in \mathcal{S}_i} (1\!-\!\langle\nabla\Phi_i(p),\bar{n}\rangle)\!\!\! \\
& + \sum\limits_{p\in\Omega} |\|\nabla\Phi_{i}(p)\|_2 - 1|  + \sum\limits_{p\in\Omega\backslash\mathcal{S}_i} \rho(\Phi_i(p)) \Big),
\end{split}
\end{equation}

\vspace{-6pt}

\noindent where $\bar{s}$ and $\bar{n}$ denote the ground-truth SDF value and surface normal respectively, $\nabla$ denotes the spacial gradient of a 3D field, 
$\Omega$ is the 3D space and $\mathcal{S}_i$ denotes shape surfaces. In practice, points will be sampled in the free space and on shape surface to calculate the loss.
The second term in Eq.~\eqref{loss:sdf} is used to learn correct normals on shape surfaces -- the gradient function of an SDF equals the surface normal given surface points as input and can be easily computed using network backpropagration.  The third term is derived from the Eikonal equation which enforces the norm of spatial gradients $\nabla \Phi_i$ to be $1$. The last term penalizes SDF values close to $0$ for non-surface points through $\rho({s})={\rm exp}(-\delta \cdot |s|), \delta \gg 1$. We refer the readers to \cite{sitzmann2020implicit} for mode details about this loss. As in \cite{park2019deepsdf}, we also apply a regularization loss to constrain the learned latent codes:
\eqvspace
\begin{equation}
L_{reg} = \sum_i\|\alpha_i\|^2_2. \label{loss:reg}
\eqvspace
\end{equation}
\noindent Alternatively, we can apply stronger regularization on the latent space
akin to the VAE training scheme~\cite{kingma2013auto}.
More details and results can be found in the \emph{suppl. material}.

\begin{figure}[t]
	\small
	\centering
	\includegraphics[width=0.97\columnwidth]{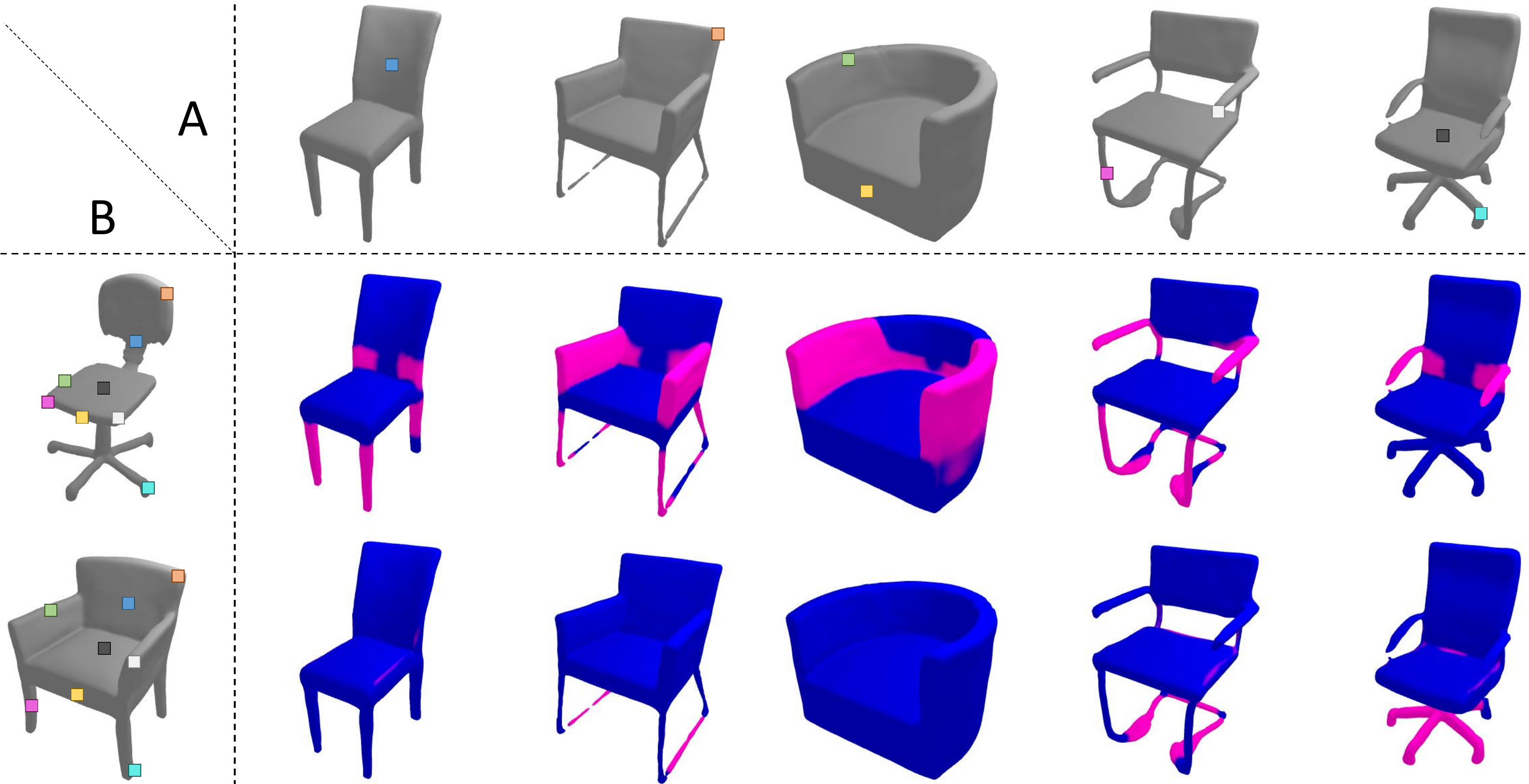}
	\vspace{-2pt}
	\caption{Correspondence uncertainty visualization. Each figure in the bottom right sector shows the uncertainty of shape A (top row)'s correspondence found on B (left column). Red and blue color represents high and low uncertainty, respectively. We also draw some colored points on A and their correspondences with the same color on B.\label{fig:uncertain}}
	\vspace{-0pt}
\end{figure}

\paravspace
\paragraph{Normal Consistency Prior.} To learn desired correspondences, our key observation is that the normal of a surface point is highly correlated with its semantic information. For example, normals on car hoods always point to the sky, and normals on the left doors always point to the left. In light of this, we encourage the normal directions of points in the template space to be consistent with their correspondences on all given shape instances:
\eqvspace
\begin{equation}
L_{normal} = \sum_i\sum\limits_{p\in\mathcal{S}_i} \Big(1-\langle\nabla T\big(p+D^v_{\omega_i}(p)\big),\bar{n}\rangle\Big), \label{eq:normal}
\end{equation}
\noindent where $\nabla T$ is the spatial gradient of template field $T$, and $\bar{n}$ denotes the ground-truth normal of point $p$ on object surface $\mathcal{S}_i$. Since the template field is used to derive all final shapes through deformation, this loss essentially enforces \emph{the normal consistency for correspondences across all the shapes} generated by our network. Note that this loss is different with the normal term in Eq.~\eqref{loss:sdf}: generating correct normals for each shape, as enforced by the latter, does not necessitate consistent correspondence normals between shapes, which is yet imposed by the former.
 
\paravspace
\paragraph{Deformation Smoothness Prior.} To encourage smooth deformation and avoid large shape distortion, we add a simple smoothness loss on the deformation field:
\eqvspace
\begin{equation}
L_{smooth} = \sum_i \sum\limits_{p\in\Omega}\sum_{d\in\{X,Y,Z\}}\|\nabla D^v_{\omega_i}|_d(p)\|_2, \label{eq:smooth}
\eqvspace
\end{equation}
\noindent which penalizes the spatial gradient of the deformation field along $X$, $Y$ and $Z$ directions.

\paravspace
\paragraph{Minimal Correction Prior.} To encourage shape representation through implicit field deformation rather than correction, we minimize the correction field via
\eqvspace
\begin{equation}
L_{c} = \sum_i\sum\limits_{p\in\Omega} | D^{\Delta s}_{\omega_i}(p)|. \label{eq:correct}
\eqvspace
\end{equation}

In summary, the whole training process can be formulated as the following optimization problem:
\eqvspace
\begin{equation}
\argmin\limits_{\{\alpha_j\},\Psi,T} L_{sdf} \!+\! w_1 L_{normal} \!+\! w_2 L_{smooth} \!+\! w_3 L_{c} \!+\! w_4 L_{reg} \label{eq:all}
\eqvspace
\end{equation}
\noindent where $w$'s are the balancing weights for different loss terms.

\begin{figure}[t]
	\small
	\centering
	\includegraphics[width=0.85\columnwidth]{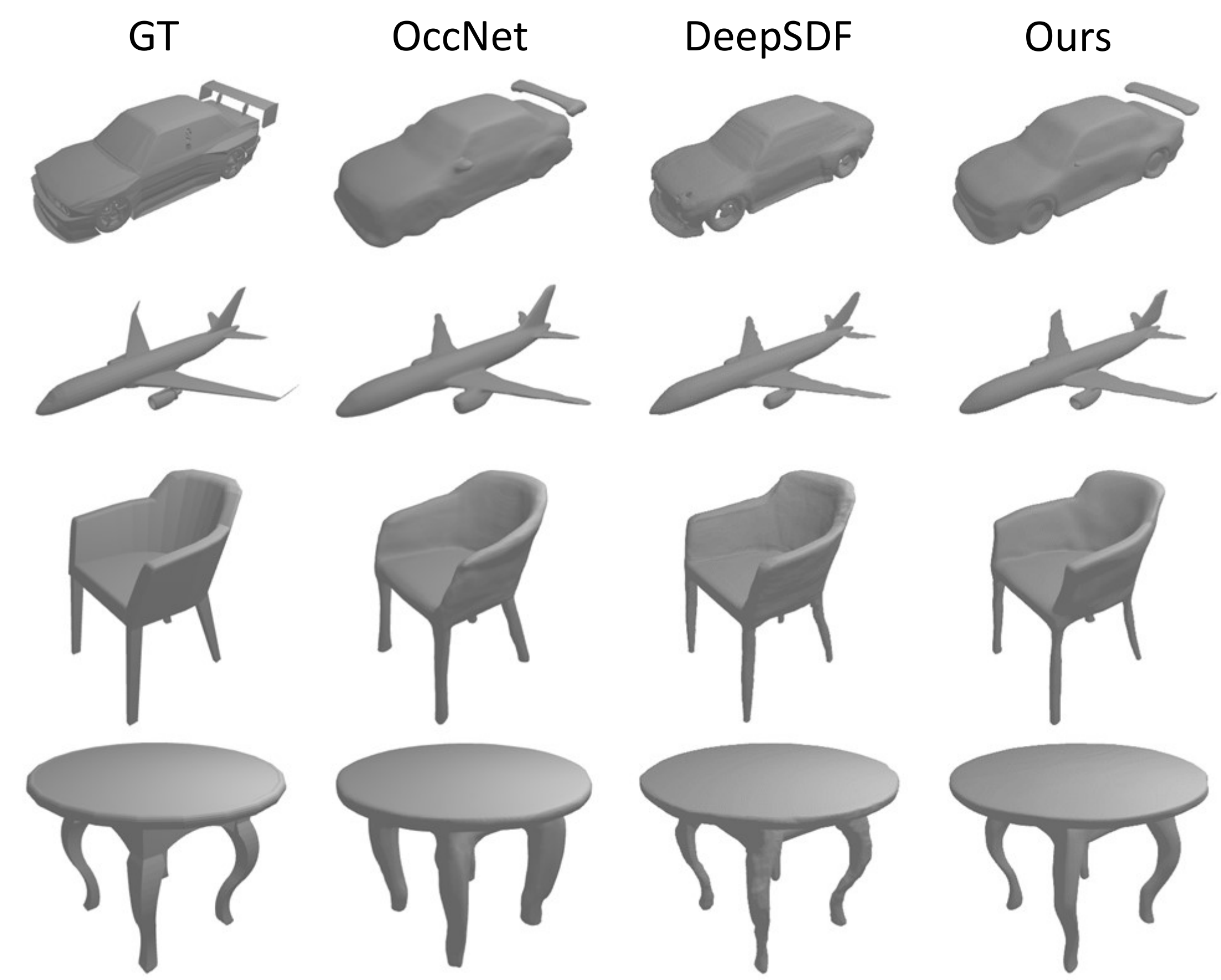}
	\caption{Shape reconstruction results for unseen shapes by OccNet~\cite{mescheder2019occupancy}, DeepSDF~\cite{park2019deepsdf} and our DIF-Net.\label{fig:test_recon}}
	\vspace{-0pt}
\end{figure}

\begin{figure*}[t]
	\small
	\centering
	\includegraphics[width=0.98\textwidth]{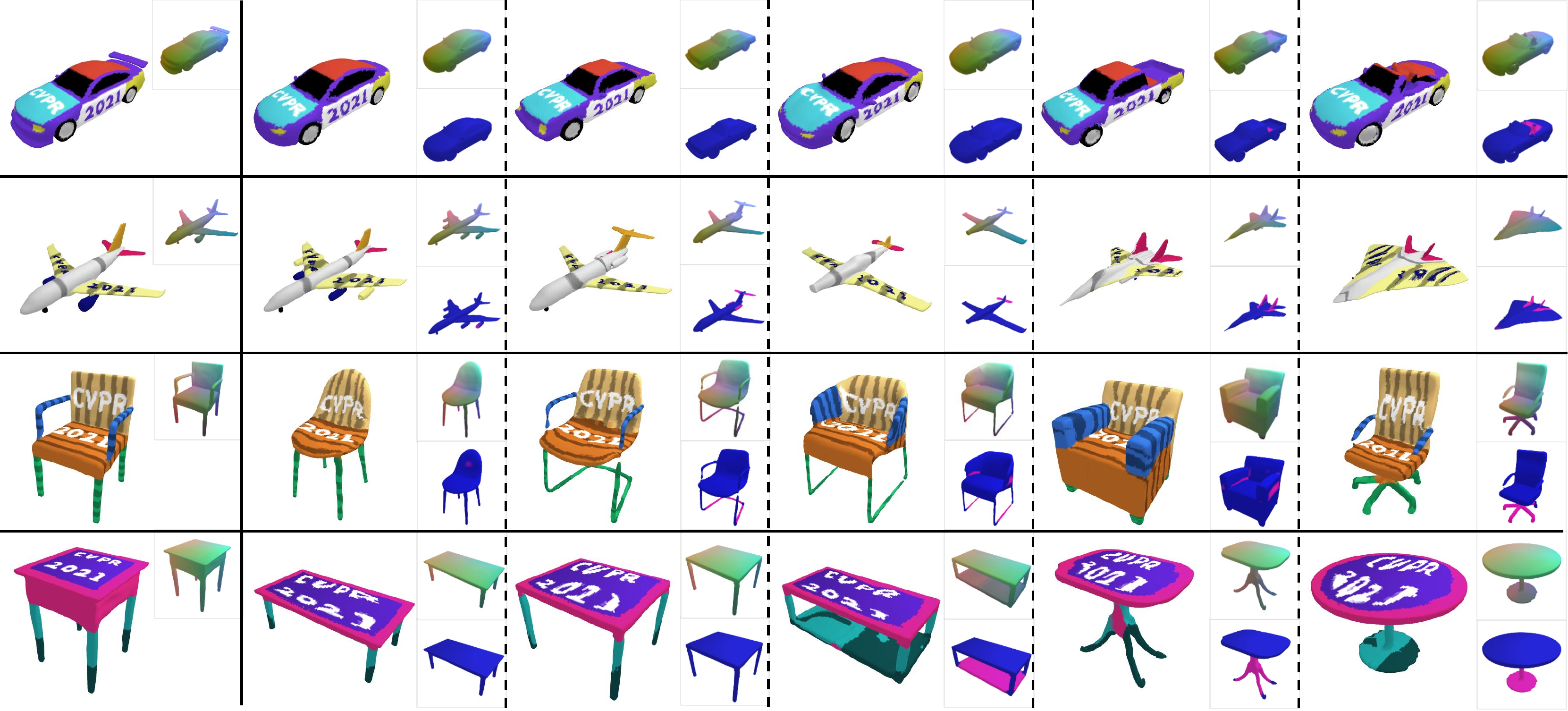}
	\caption{Qualitative evaluation of our learned dense correspondence for each category. For better visualization, we manually paint a generated shape (first column) with different colors on different semantic components. We also draw some strip patterns and texts on the shapes. Then we transfer these colors to other shapes generated by our method (last five columns) according to their correspondences. We also visualize the correspondences color-coded by spatial coordinates as well as the correspondence uncertainty for each shape. \label{fig:main}
	}
	\vspace{-5pt}
\end{figure*}

\subsection{Correspondence Uncertainty Measurement}

In practice, it is desirable to have a quality or uncertainty metric for the obtained shape correspondences, which can be used for structure difference analysis, bad correspondence removal \etc. As mentioned previously, the correspondence between two objects $\mathcal{O}_i$ and $\mathcal{O}_j$ can be built by nearest neighbor search in the template space. 
Let $p_i$ be a point on $\mathcal{O}_i$ and $p_j$ its corresponding points found on $\mathcal{O}_j$, we propose a simple yet surprisingly-effective uncertainty metric based on their distance in the template space:
\eqvspace
\begin{equation}
u(p_i,p_j) = 1 - {\rm exp} (-\gamma\|(p_i+v_i)-(p_j+v_j)\|_2^2) \label{eq:uncertain}
\eqvspace
\end{equation} 
\noindent where $v_i = D^v_{\omega_i}(p_i)$ is the deformation vector (similarly for $v_j$) and $\gamma$ is a scaling factor. The examples in Fig.~\ref{fig:uncertain} show that the regions with high uncertainty computed by Eq.~\eqref{eq:uncertain} conform well to structure discrepancy between shapes.

\section{Experiments}

\paragraph{Implementation Details.}
The Hyper-Net $\Psi$, Deform-Net $D$ and template field network $T$ are all implemented as MLPs. We train them on four categories in ShapeNet-V2~\cite{chang2015shapenet}, including \emph{car}, \emph{airplane}, \emph{chair}, and \emph{table}. 
All parameters are trained end-to-end using the Adam \cite{kingma2014adam} optimizer.
Training takes about 4 hours on 8 NVIDIA V100 GPUs with batchsize 256 for one category. See \emph{suppl. material} for more details.

\subsection{Ability of Shape Representation}
To evaluate the representation power of DIF and the latent shape space learned by DIF-Net, we embed new shapes \emph{unseen} in the training stage and measure the reconstruction accuracy. With our trained networks $\Psi$ and $T$, we embed a test shape to the latent space by solving the following simplified optimization problem of Eq.~\eqref{eq:all}:
\eqvspace
\begin{equation}
\argmin\limits_{\alpha} L_{sdf} + w_4 L_{reg}. \label{eq:test_recon}
\eqvspace
\end{equation}
\noindent We compare with two state-of-the-art shape modeling methods based on deep implicit field: OccNet \cite{mescheder2019occupancy} and DeepSDF \cite{park2019deepsdf}. For OccNet, we train an individual model for each category using our training data for a fair comparison. For DeepSDF, we use a per-category model trained by \cite{liu2020dist} to evaluate its performance.

\begin{table}[t]
	\centering
	\small
	\begin{tabular}{c|c|c|c|c}
		\hline
		CD ($\times1000$) &car&plane&chair&table \\
		\hline
		OccNet* \cite{mescheder2019occupancy}& 0.582 & 0.288 & 0.995 & 1.326 \\
		DeepSDF$^\dag$ \cite{park2019deepsdf} & 0.767 & 0.298& 0.785& 1.422\\
		Ours& \textbf{0.404} & \textbf{0.249} & \textbf{0.661} & \textbf{1.036} \\
		\hline
		Ours w/o deform. & 0.353 & 0.255 & 0.529 & 0.772\\
		\hline
		\hline
		EMD &car&plane&chair&table \\
		\hline
		OccNet* \cite{mescheder2019occupancy}& 0.037 & 0.025 & 0.045 & 0.047 \\
		DeepSDF$^\dag$ \cite{park2019deepsdf} & 0.041 & 0.029 & 0.038 & 0.046\\
		Ours& \textbf{0.036} & \textbf{0.024}& \textbf{0.038} & \textbf{0.040} \\
		\hline
		Ours w/o deform. & 0.034& 0.025& 0.036 & 0.037\\
		\hline
	\end{tabular}
	\vspace{3pt}
	\caption{Reconstruction accuracy for unseen shapes. We use the first 100 shapes in the intersection of the test set splits from DeepSDF \cite{park2019deepsdf} and ours. OccNet~\cite{mescheder2019occupancy} is trained for each category using our training data, and a per-category DeepSDF model trained by \cite{liu2020dist} is evaluated here. Reconstructed meshes are extracted at a resolution of $256^3$ for all methods. CD and EMD are evaluated using 10K and 8K sampled points respectively.\label{tab:represent}}
	\vspace{-5pt}
\end{table}

Table~\ref{tab:represent} shows the shape reconstruction accuracy on 100 test shapes for each category, measured with chamfer distance (CD) and earth mover distance (EMD), and Fig.~\ref{fig:test_recon} visually compares some results. It can be seen that all three methods perform well in representing unseen shapes, and our method is slightly better in terms of numerical error.

We also compare with a variant of our method which does not model dense correspondence. Specifically, we replace DIF-Net $\Phi$ with a MLP of three hidden layers that directly predicts the SDF of a shape. The numerical results are presented in Table~\ref{tab:represent}, which are slightly better than our DIF-Net. It indicates that the deformation-based implicit design only leads to moderate decease of representation capability. However, high-quality dense correspondences can be achieved with this design, as we will show later. 

Due to space limitation, more evaluations of our trained DIF-Net including \textbf{latent space interpolation}, \textbf{sampling} and \textbf{retrieval} are deferred to the \emph{suppl. materials}.

\subsection{Learned Dense Correspondence}\label{sec:exp_correspondence}

\paragraph{Qualitative Evaluation.}
Figure~\ref{fig:main} visualizes the correspondences generated by our method, where we manually paint salient color patterns on the shapes to better check the correspondence quality. Visually inspected, our method produces convincing correspondences across various shapes despite their structure differences. It not only correctly matches shared semantic components between two shapes but also preserves the original color patterns. Moreover, the uncertain regions revealed by our method well reflect structure differences between two shapes.

\begin{figure}[t]
	\small
	\centering
	\includegraphics[width=0.97\columnwidth]{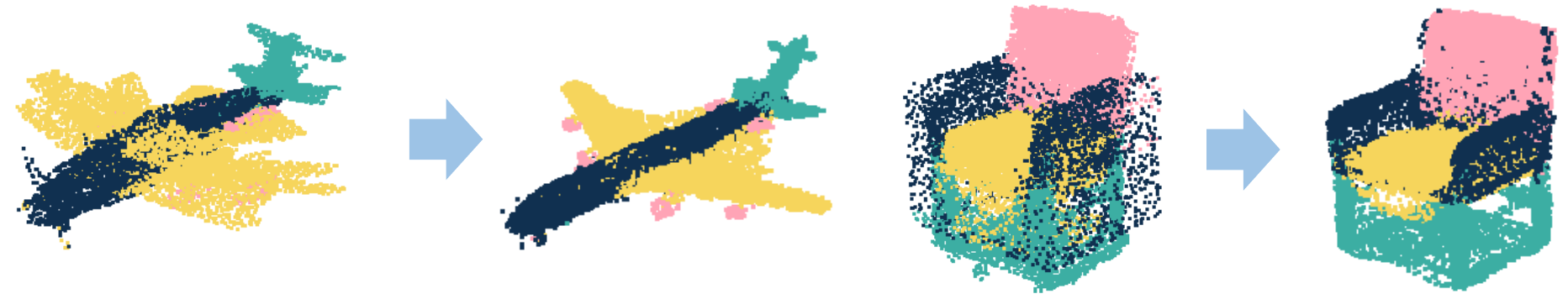}
	\caption{Labeled points on the original shapes are deformed into the template 3D space where the semantic parts are well aligned \label{fig:seg_deform}}
	\vspace{0pt}
\end{figure}

\begin{table}[t]
	\centering
	\small
	\footnotesize
	\begin{tabular}{c|c|c|c|c|c}
		\hline
		IoU &car&plane&chair&table&average \\
		\hline
		\!\!\!Closest Point\!\!\! & \!\!\!62.7\,$|$\,63.5\!\!\! & \!\!\!60.5\,$|$\,62.5\!\!\! & \!\!\!65.9\,$|$\,69.6\!\!\! & \!\!\!68.5\,$|$\,73.4\!\!\! & \!\!\!64.4\,$|$\,67.2\!\!\! \\
		\!\!\!\!Atlas-sph.*\,\cite{groueix2018papier}\!\!\!\!  & \!\!\!62.6\,$|$\,64.0\!\!\! & \!\!\!51.1\,$|$\,50.9\!\!\! & \!\!\!56.7\,$|$\,57.9\!\!\! & \!\!\!64.0\,$|$\,67.0\!\!\! & \!\!\!58.6\,$|$\,59.9\!\!\! \\
		\!\!Atlas-25*~\cite{groueix2018papier}\!\! & \!\!\!59.3\,$|$\,60.3\!\!\! & \!\!\!54.2\,$|$\,52.2\!\!\! & \!\!\!62.1\,$|$\,64.9\!\!\! & \!\!\!66.2\,$|$\,69.1\!\!\! & \!\!\!60.5\,$|$\,61.6\!\!\! \\
		\!\!Atlas-v2*~\cite{deprelle2019learning}\!\! & \!\!\!64.8\,$|$\,65.9\!\!\! & \!\!\!53.2\,$|$\,52.6\!\!\! & \!\!\!63.4\,$|$\,65.7\!\!\! & \!\!\!65.6\,$|$\,68.0\!\!\! & \!\!\!61.8\,$|$\,63.0\!\!\! \\
		SIF* \cite{genova2019learning}& \!\!\!62.6\,$|$\,63.9\!\!\!& \!\!\!52.3\,$|$\,52.3\!\!\! &\!\!\!57.6\,$|$\,57.0\!\!\! & \!\!\!65.7\,$|$\,68.7\!\!\! & \!\!\!59.6\,$|$\,60.5\!\!\!\\
		Ours* & \!\!\!\textbf{72.7}\,$|$\,\textbf{74.1}\!\!\!&\!\!\!\textbf{71.7}\,$|$\,\textbf{78.4}\!\!\! & \!\!\!\textbf{75.3}\,$|$\,\textbf{79.7}\!\!\! & \!\!\!\textbf{81.1}\,$|$\,\textbf{87.9}\!\!\!& \!\!\!\textbf{75.2}\,$|$\,\textbf{80.0}\!\!\!\\
		\hline
		\!\!\!\!\!ShapeFlow$^\dag$ \cite{jiang2020shapeflow}\!\!\!\!\!& \!\!\!-\!\!\!& \!\!\!-\!\!\! &\!\!\!71.6\,$|$\,77.5\!\!\! & \!\!\!-\!\!\! & \!\!\!-\!\!\!\\
		\!\!\!\!DualSDF$^\dag$ \cite{hao2020dualsdf}\!\!\!\!& \!\!\!-\!\!\!& \!\!\!54.9\,$|$\,53.5\!\!\! &\!\!\!58.1\,$|$\,59.1\!\!\! & \!\!\!-\!\!\! & \!\!\!56.5\,$|$\,56.3\!\!\!\\
		Ours$^\dag$ & \!\!\!-\!\!\!&\!\!\!\textbf{68.8}\,$|$\,\textbf{74.7}\!\!\! & \!\!\!\textbf{75.3}\,$|$\,\textbf{80.9}\!\!\! & \!\!\!-\!\!\!& \!\!\!\textbf{72.0}\,$|$\,\textbf{77.8}\!\!\!\\
		\hline
	
		\hline
	\end{tabular}
	\vspace{0pt}
	\caption{Label IoU (mean$|$median) on label transfer task. For each category, we use 5 labeled source shapes and test the segmentation accuracy on our whole training set containing 3K-4K shapes. *: Trained and tested for each category individually using our training set. $^\dag$: Trained by original papers and tested on all shapes in the intersection of our training set and theirs.
		\label{tab:corres}}
		\vspace{-4pt}
\end{table}

\paravspace
\vspace{-1pt}
\paragraph{Quantitative Evaluation via Label Transfer.} To our knowledge, there is no dataset offering ground truth dense correspondence for objects with structure variation. Therefore, we resort to a semantic label transfer experiment for quantitative evaluation. We use the ShapeNet-Part dataset~\cite{yi2016scalable} which contains part labels for ShapeNet objects. For each of the four object categories, we selected 5 labeled shapes as source shapes, and transfer their labels to other shapes leveraging dense correspondences. This task can be viewed as few-shot 3D shape segmentation learning using 5 samples as training data.

\begin{figure}[t]
	\small
	\centering
	\includegraphics[width=1.0\columnwidth]{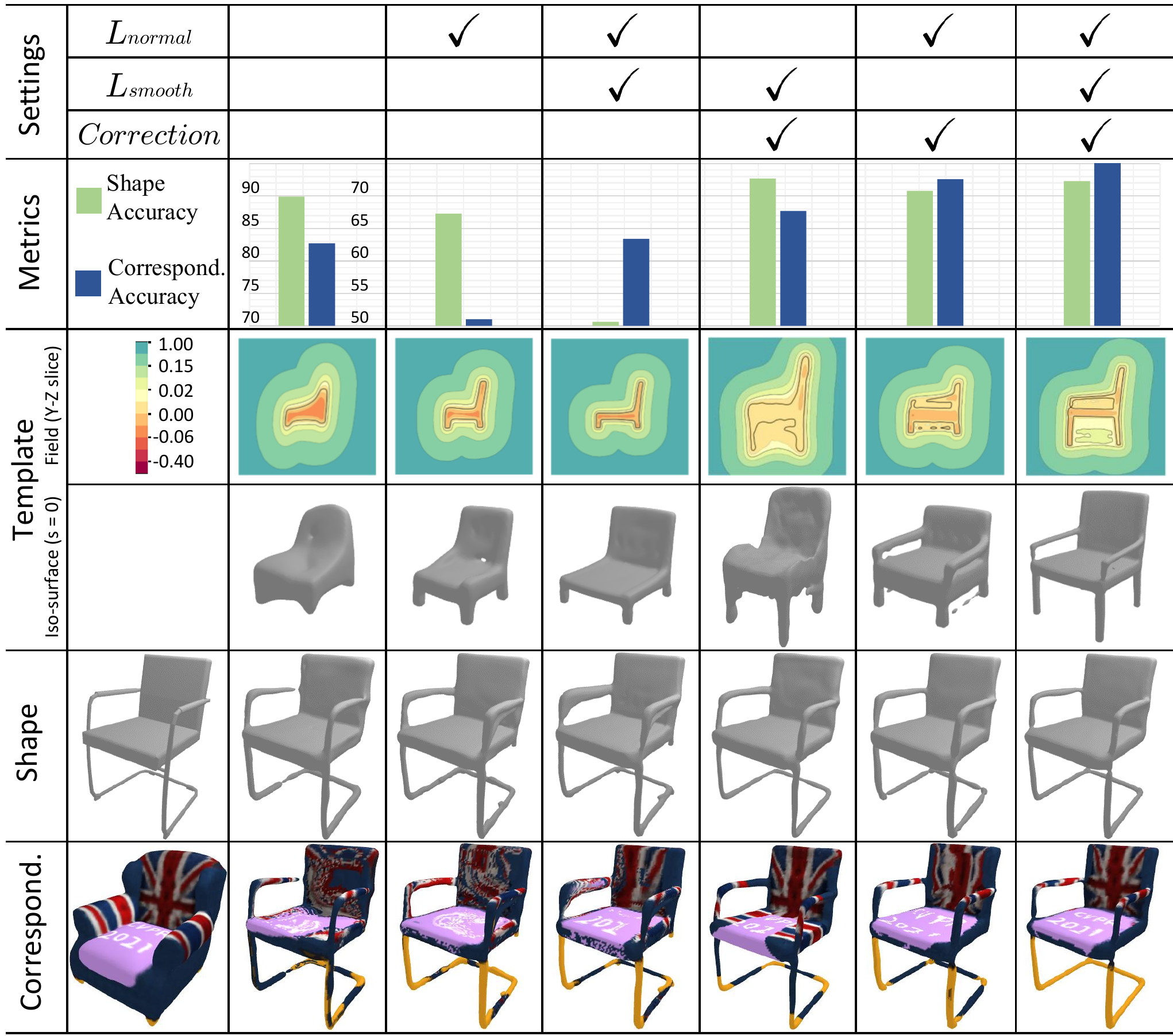}
	\caption{Influence of different training losses and the correction field in our model. We use the chair category for evaluation, and measure shape reconstruction accuracy via F-score~\cite{tatarchenko2019single} at $\tau=0.001$ and correspondence accuracy via label IoU in the part label transfer task of Section~\ref{sec:exp_correspondence}. We also present visual results to further illustrate the effect of different components.\label{fig:ablation}}
	\vspace{-2pt}
\end{figure}

For this task, we first deform all labeled points on the 5 source shapes into our template 3D space, as shown in Fig.~\ref{fig:seg_deform}. 
For an unlabeled shape, we deform its surface points into the template 3D space, find 10 nearest labeled points for each, and then conduct simple label voting. 

We compare our method with AtlasNet~\cite{groueix2018papier}, AtlasNet-v2~\cite{deprelle2019learning}, SIF~\cite{genova2019learning}, ShapeFlow~\cite{jiang2020shapeflow}, and DualSDF~\cite{hao2020dualsdf}, which can be used to build correspondences, as well as a naive closest point based method. 
For fair comparison, we train AtlasNet, AtlasNet-v2, and SIF for each category individually with our training data.
For Shapeflow and DualSDF, we use a per-category model trained by the original papers and use all shapes in the intersection of our training set and theirs as target shapes. More details and the visual results can be found in the \emph{suppl. material}.

Table~\ref{tab:corres} compares the accuracy measured by IoU between ground-truth and transferred labels. Our method outperforms all others by a wide margin. 
AtlasNet, AtlasNet-v2, DualSDF, and SIF may generate inconsistent correspondences (\eg, a point labeled as chair ``arm" corresponds to ``arm" regions for some shapes but to ``seat" for others) due to the limitation of their shape representations. Their results are worse than the naive closest point method for categories with large shape variation. ShapeFlow alleviates this problem by learning a volumetric deformation flow, but still suffers from inaccurate correspondence reasoning due to a mesh-based shape representation.

\begin{figure*}[t]
	\small
	\centering
	\includegraphics[width=0.98\textwidth]{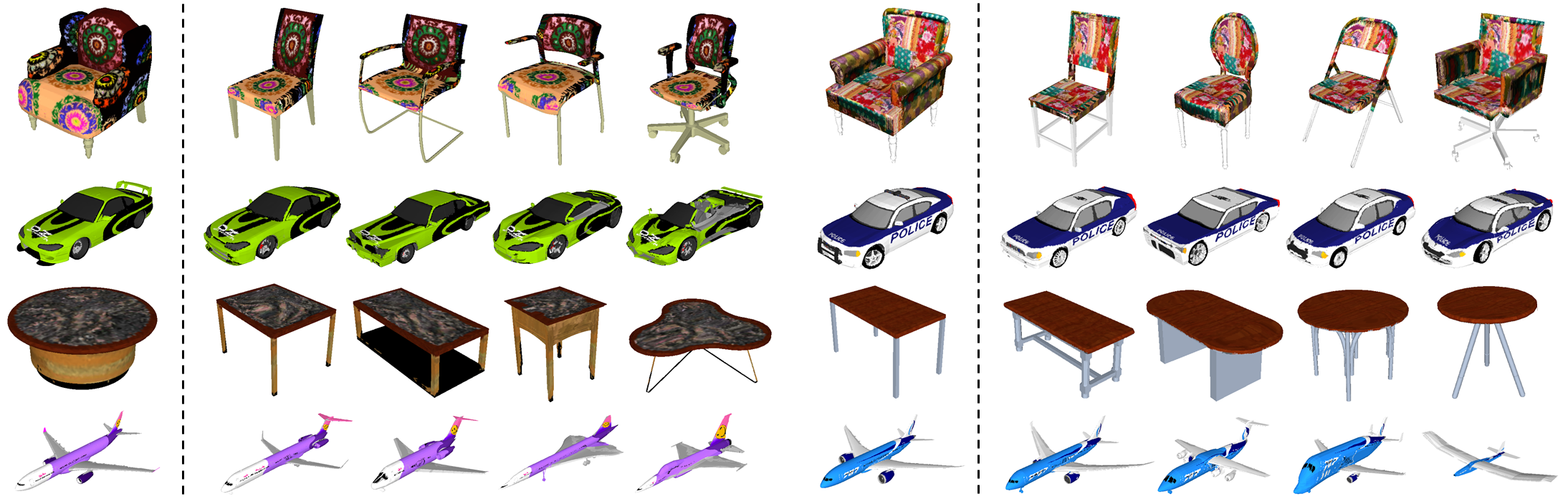}
	\caption{Texture transfer result on ShapeNet objects using correspondences generated by our DIF-Net. (\textbf{Best viewed with zoom})\label{fig:transfer}}
	\vspace{-8pt}
\end{figure*}

\begin{figure}[t]
	\small
	\centering
	\includegraphics[width=0.98\columnwidth]{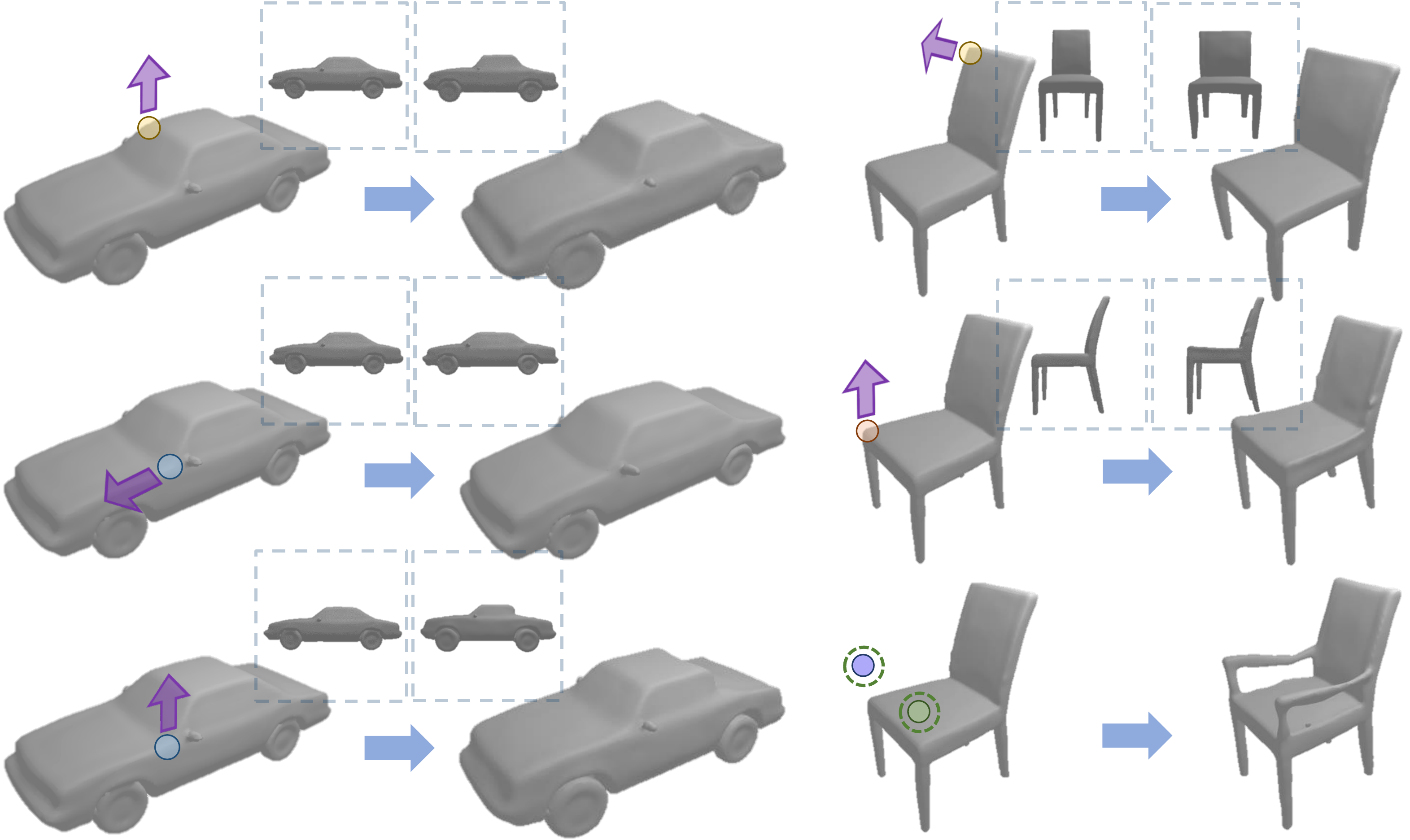}
	\caption{Shape editing result. DIF-Net can deform shapes and add new structures using only sparse points as guidance.\label{fig:edit}}
	\vspace{-7pt}
\end{figure}

\subsection{Ablation Study} \label{sec:ablation}
In this section,  we conduct ablation study to validate the efficacy of our training loss terms and the correction field. 
The main results of different settings are shown in Fig.~\ref{fig:ablation}.
As can be seen, without the normal loss $L_{normal}$, the learned correspondences are inferior as indicated by the significant IoU drop in label transfer. Without the deformation smoothness loss $L_{smooth}$, the learned correspondences are highly distorted, as shown in the transferred textures. Without the correction field, shape representation ability of the model decreases significantly.
Figure~\ref{fig:ablation} shows that the correction field is also crucial to obtain high quality correspondences. 
Without the correction field, more complex deformations are needed to represent the final shape with various structures, which significantly increases the learning difficulty. Under this situation, learning may get stuck into local minima, leading to an inferior template field lacking rich structural information (as shown by the iso-surfaces in Fig.~\ref{fig:ablation}), which further results in correspondence accuracy drop. 

\section{Applications}

\subsection{Texture Transfer}
Using dense correspondence generated by DIF-Net, we are able to transfer textures from one object to another. Successfully transferring rich textures among various shapes necessitates high quality dense correspondence. Figure~\ref{fig:transfer} shows texture transfer results between ground truth shapes in ShapeNet. Visually inspected, the rich texture patterns are well preserved and transferred to correct semantic areas in new shapes. Figure~\ref{fig:teaser} contains two more texture transfer results of our method for embedded 3D shapes.

\subsection{Shape Editing}
With the learned latent shape and dense correspondence, our method can be used to manipulate 3D shapes by moving one or a sparse set of points. Specifically, give a shape with embedded latent code $\alpha$,  we can freely select one 3D point $p_1$ on the shape and specify its desired new position $p_2$. Let $p_1'=p_1 + D_{\Psi({\alpha})}^v(p_1)$ be the deformed point of $p_1$ in the template 3D space, we achieve shape editing via solving for a new shape code $\hat{\alpha}$ minimizing the following equation:
\eqvspace
\begin{equation}
\argmin\limits_{\hat{\alpha}} \|(p_2+v)-p_1'\|_2^2 + |\Phi_{\Psi(\hat{\alpha})}(p_2)| + \|\hat{\alpha}-\alpha\|_1 
\label{eq:edit}
\eqvspace
\end{equation}
\noindent where $v=D_{\Psi(\hat{\alpha})}^v(p_2)$ is the deformation vector for $p_2$ with the new shape code $\hat{\alpha}$.
In this equation, the first term enforces the original and new points on the shapes before and after editing to be a correspondence pair thus having same semantic meaning. The second term ensures the new point lies on the new shape surface. The third term requires the code change to be small. Figure~\ref{fig:edit} shows the editing results for two shapes and another two examples can be found in Fig.~\ref{fig:teaser}. We can even add new structures to a given shape via Eq.~\eqref{eq:edit}, where in this case we directly select $p_1'$ in the template space and a free point $p_2$ in the shape space. An example is shown in Fig.~\ref{fig:edit}. More details and results regarding shape editing can be found in the \emph{suppl. material}.

\section{Conclusion}
We have presented Deformed Implicit Field, a novel implicit-based representation modeling a class of 3D shapes and providing dense correspondences. We also presented DIF-Net, a neural shape model that learns high-quality dense correspondences in an unsupervised manner through our proposed loss functions. 
Various experiments and applications collectively demonstrated the high quality shapes and correspondences generated by our method. In future, we plan to extend the DIF representation to handling more generic 3D objects and scenes. 

{\small
	\bibliographystyle{ieee_fullname}
	\bibliography{DIF_NET}
}

\clearpage

\appendix

\renewcommand{\thesection}{\Alph{section}}
\renewcommand{\thefigure}{\Roman{figure}}
\renewcommand{\thetable}{\Roman{table}}
\renewcommand{\theequation}{\Roman{equation}}
\setcounter{figure}{0}
\setcounter{equation}{0}

\section{More Implementation Details}

\paragraph{Data Preparation.} 
To train DIF-Net via the SDF regression loss $L_{sdf}$ defined in Eq.~(\ref{loss:sdf}) in the main paper, we randomly sample points on shape surface and in the free space.
Specifically, we follow a similar step as in \cite{park2019deepsdf} to normalize each ground truth mesh into a sphere with radius of 1/1.03. Then, for surface points, we render 100 virtual images for each normalized mesh using 100 virtual cameras regularly sampled on the unit sphere. Surface points are obtained via back-projecting the depth pixels from these virtual images. Normals of these surface points are obtained in a similar way from virtual normal images. This sampling strategy helps us get rid of inner structures of each mesh that are invisible from outside (e.g car seats).

For free-space points, we uniformly sample them within a cube of $[-1,1]^3$, and calculate their distance to the nearest surface points sampled using the above strategy. To decide the sign of distance for a free-space point, we render it with the same virtual cameras used to obtain surface points, and check if its depth is smaller than the depth of a surface point falling into the same pixel. As long as it has a smaller depth value in any virtual image, it will be classified as an external point and get a positive sign. Otherwise, it will be classified as an internal point and get a negative sign.

In the end, we randomly sample 500K surface points along with normals and 500K free space points with their SDF values for each mesh.

\begin{figure}[t]
	\small
	\centering
	\includegraphics[width=1.0\columnwidth]{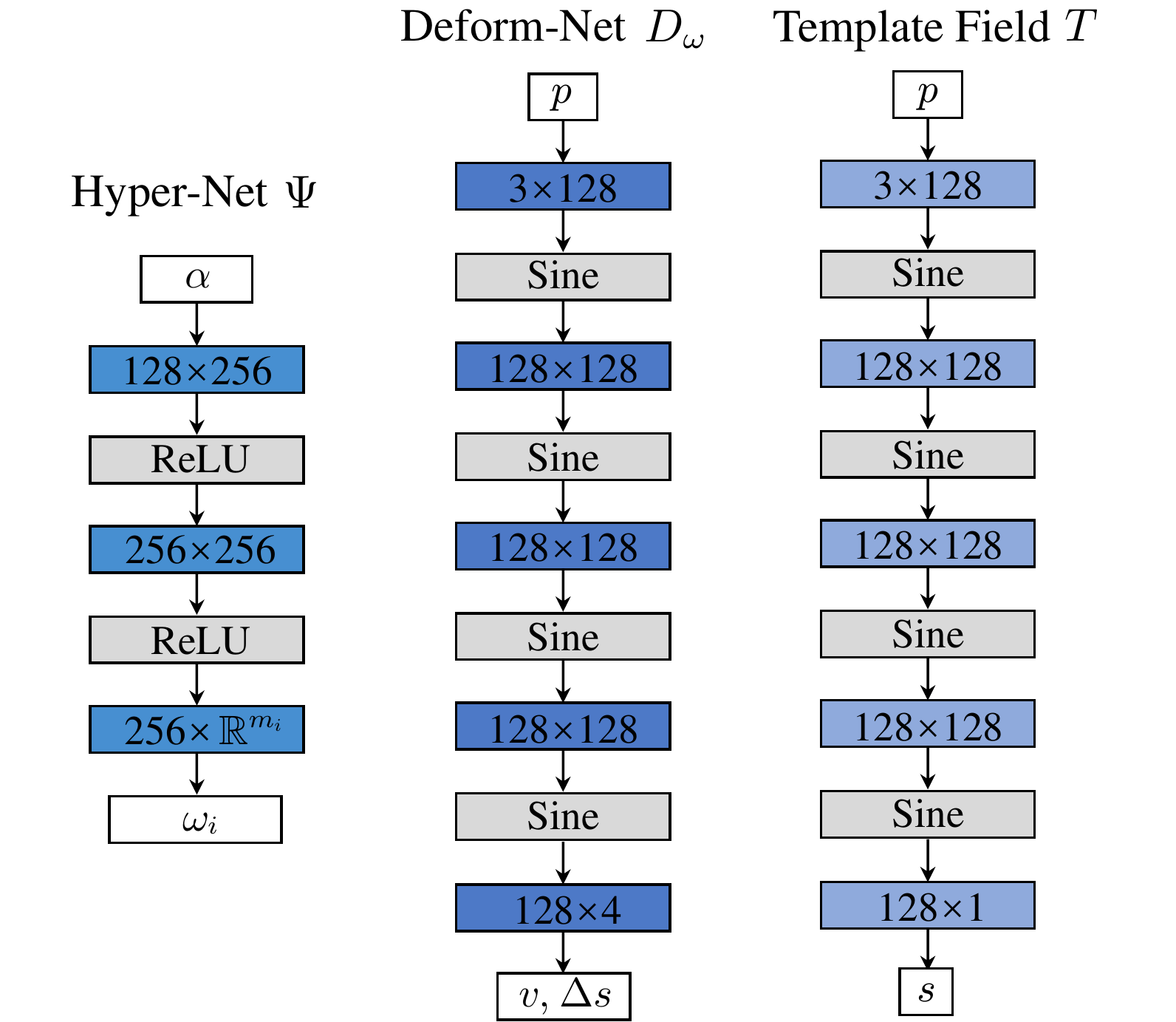}
	\caption{Structures of different networks used in our method.
	\label{fig:net}}
	\vspace{-5pt}
\end{figure}

\paravspace
\paragraph{Network Architecture.} Figure~\ref{fig:net} shows the structures of Hyper-Net $\Psi$ and DIF-Net $\Phi$ used in our method. All networks are MLPs. The Hyper-Net consists of multiple MLPs each responsible for the weights $\omega_i$ of a single fully-connected layer $i$ in the Deform-Net $D_\omega$. The weights of Template Field $T$ are shared across the class. For Hyper-Net $\Psi$, we use ReLU activations. For Deform-Net $D_\omega$ and Template Field $T$, we use sine activations advocated by \cite{sitzmann2020implicit}.

\paravspace
\paragraph{Training Details.} We train our model on four categories in ShapeNet-V2~\cite{chang2015shapenet}, including \emph{car}, \emph{airplane}, \emph{chair}, and \emph{table} using 3K, 3.5K, 4K, and 4K instances as training set, respectively.

We jointly learn all latent codes $\{\alpha_i\}$, the Hyper-Net $\Psi$, and the Template Field $T$ using the losses in Eq.~(\ref{eq:all}) in the main paper. Similar to~\cite{park2019deepsdf}, we initialize all latent codes using $\mathcal{N}(0,0.01^2)$. The weights of Hyper-Net $\Psi$ are initialized using~\cite{he2015delving}. The weights of Template Field $T$ are initialized as in~\cite{sitzmann2020implicit}.

We train an individual model on each category for 60 epochs. At each iteration, we randomly select 4K surface points and 4K free-space points for each shape in a batch to calculate the losses in Eq.~(\ref{eq:all}) and update learnable parameters. During each epoch, there are in total 200K surface points and 200K free-space points used for each shape. 

We set the balancing weights for four terms in the SDF regression loss $L_{sdf}$ in Eq.~(\ref{loss:sdf}) to $3e3$, $1e2$, $5e1$, and $5e2$ respectively following~\cite{sitzmann2020implicit}. $w_1$ and $w_4$ are set to $1e2$ and $1e6$ respectively for all categories. For $w_2$, we set it to $5$, $2$, $5$, and $1$ for \emph{car}, \emph{airplane}, \emph{chair}, and \emph{table} respectively. For $w_3$, we set it to $1e2$, $1e2$, $5e1$, and $1e2$ for each of the above categories, respectively. Models for all categories are trained using an Adam optimizer \cite{kingma2014adam} with a learning rate of $1e-4$ and a batchsize of 256.

\paravspace
\paragraph{Inference Details.} At inference time, given a test shape, we obtain its latent code via optimizing Eq.~(\ref{eq:test_recon}). Weights for different loss terms and point sampling strategy are identical to the training phase. We use Adam optimizer with a learning rate of $1e-4$, and update the latent code for 30 epochs in total.

\section{Different Latent Code Regularization}
In the main paper, we constrain the norm of learned latent code $\alpha$ to be small using $L_{reg}$ defined in Eq.~(\ref{loss:reg}). Alternatively, we can also constrain the posterior distribution of latent code to be close to a Gaussian distribution to learn a latent space for better sampling. Specifically, we can replace the $L_{reg}$ in Eq.~(\ref{loss:reg}) with the following loss:
\eqvspace
\begin{equation}
L_{reg'} = KL(q(\alpha_i|\mathcal{O}_i)\|\mathcal{N}(\mu,\Sigma^\intercal\Sigma)) \label{loss:reg2}
\eqvspace
\end{equation}
where $KL$ denotes the Kullback--Leibler divergence and $q(\alpha_i|\mathcal{O}_i)$ is the posterior distribution of $\alpha_i$ represented by a Gaussian distribution with mean equals to $\alpha_i$. To calculate $L_{reg'}$ in Eq.~(\ref{loss:reg2}), we have to obtain the standard deviation for $q(\alpha_i|\mathcal{O}_i)$ as well. Therefore, we introduce an extra learnable latent code $\sigma$ with the same dimension of $\alpha$ to represent its standard deviation.

Specifically, at training stage, the input latent code $\alpha$ to the Hyper-Net $\Psi$ is replaced by a random variable $\widetilde{\alpha}$ sampled from $\mathcal{N}(\alpha,\sigma^2I)$. Then, we train all learnable parameters using the following losses:
\eqvspace
\begin{equation}
\argmin\limits_{\{\alpha_j\},\{\sigma_j\},\Psi,T} \! L_{sdf} \!+\! w_1 L_{normal} \!+\! w_2 L_{smooth} \!+\! w_3 L_{c} \!+\! w_5 L_{reg'}, \label{eq:all2}
\eqvspace
\end{equation}
where $L_{reg'}$ is defined in Eq.~(\ref{loss:reg2}) and all other losses are the same as in Eq.~(\ref{eq:all}).
In practice, we set the Gaussian distribution in Eq.~(\ref{loss:reg2}) to $\mathcal{N}(0,0.01^2)$, and set the weight $w_5$ to $1e2$. Other balancing weights are the same as described in the previous section. 

\begin{figure}[t]
	\small
	\centering
	\includegraphics[width=1.0\columnwidth]{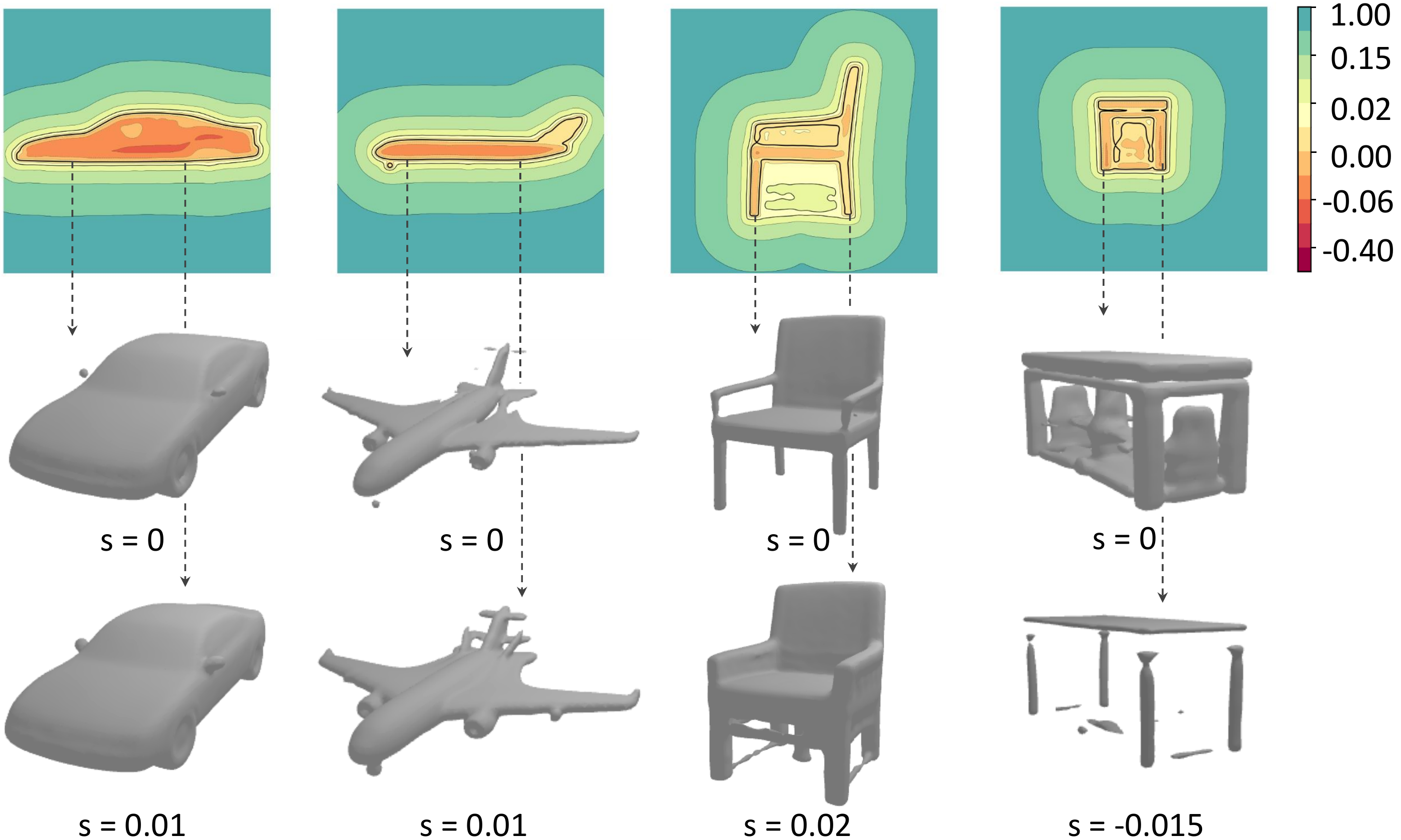}
	\caption{The learned template implicit field (\emph{Y-Z} slice) for four categories and different iso-surfaces extracted from them.
	\label{fig:mean_all}}
	\vspace{-2pt}
\end{figure}

\section{Learned Template Field}

Figure~\ref{fig:mean_all} shows the learned template fields for four categories. Different structures of a category are recorded in different iso-surfaces of the learned template field.

\section{Learned Shape Latent Space}
\begin{figure*}[t]
	\small
	\centering
	\includegraphics[width=0.97\textwidth]{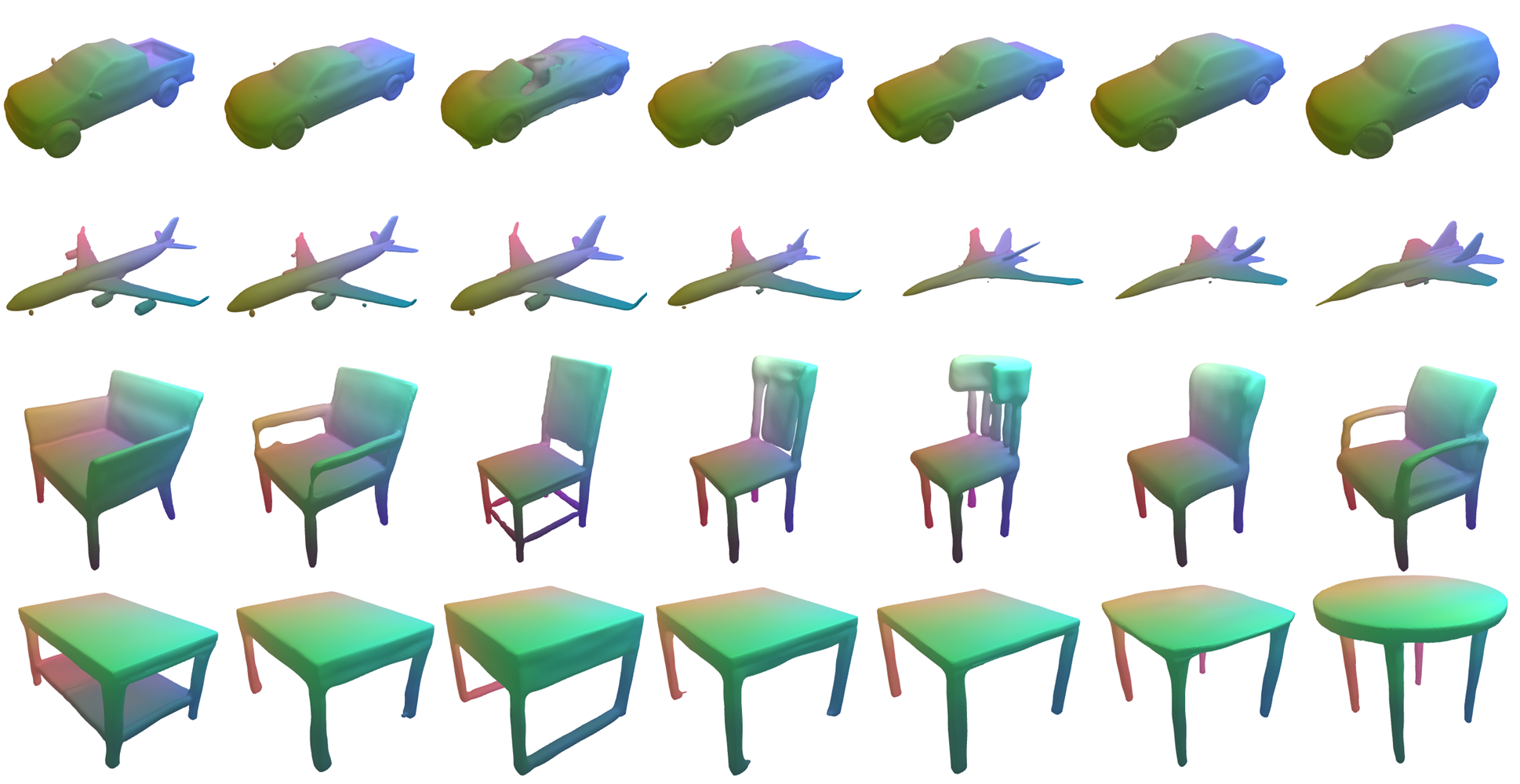}
	\caption{Latent code interpolation results. Interpolated shapes are in even columns.
	\label{fig:interp}}
\end{figure*}

\paragraph{Latent Space Interpolation.} In this part, we show latent space interpolation results using the model trained with Eq.~(\ref{eq:all}) in Fig.~\ref{fig:interp}. Shapes in even positions are interpolated from their two neighbors. We also show the color-coded correspondence of them. As depicted, interpolated shapes are reasonable and correspondence between different shapes are consistent.

\begin{figure*}[t]
	\small
	\centering
	\includegraphics[width=0.97\textwidth]{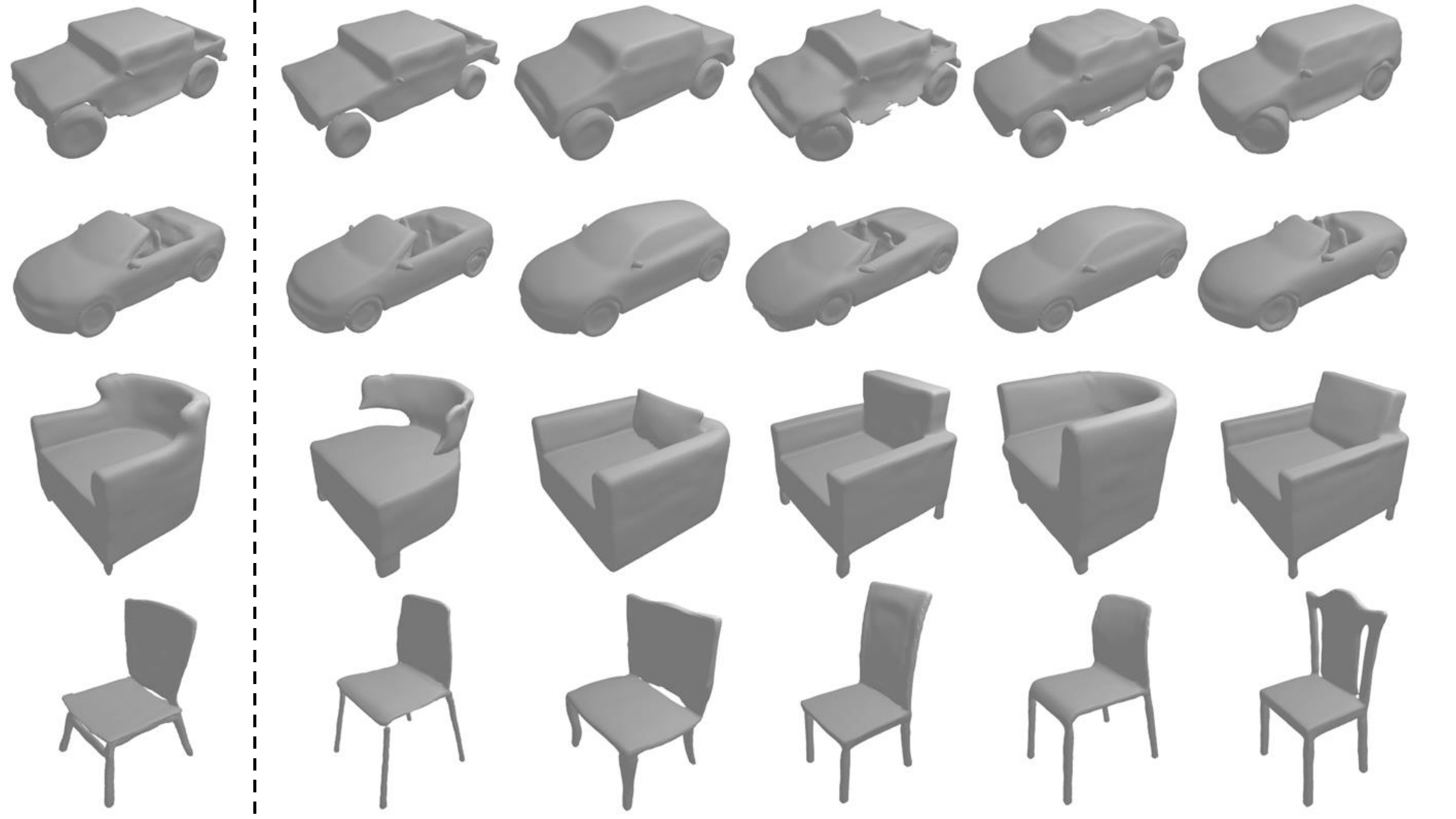}
	\caption{Shape retrieval results using shape latent codes. Shapes on the right-hand side are nearest neighbors of the left-most ones and are sorted in an ascending distance order.
	\label{fig:retri}}
\end{figure*}

\paravspace
\paragraph{Latent Space Retrieval.} Given a source shape, we can search for its nearest neighbors using the Euclidean distance between shape latent codes as metrics. Retrieval results are shown in Fig.~\ref{fig:retri}. Shapes on the right-hand side are nearest neighbors of the left-most shapes, and are sorted in an ascending distance order. Our learned latent space can capture the shape similarities in the original shape space---similar shapes are embeded close to each other.

\begin{figure*}[t]
	\small
	\centering
	\includegraphics[width=0.97\textwidth]{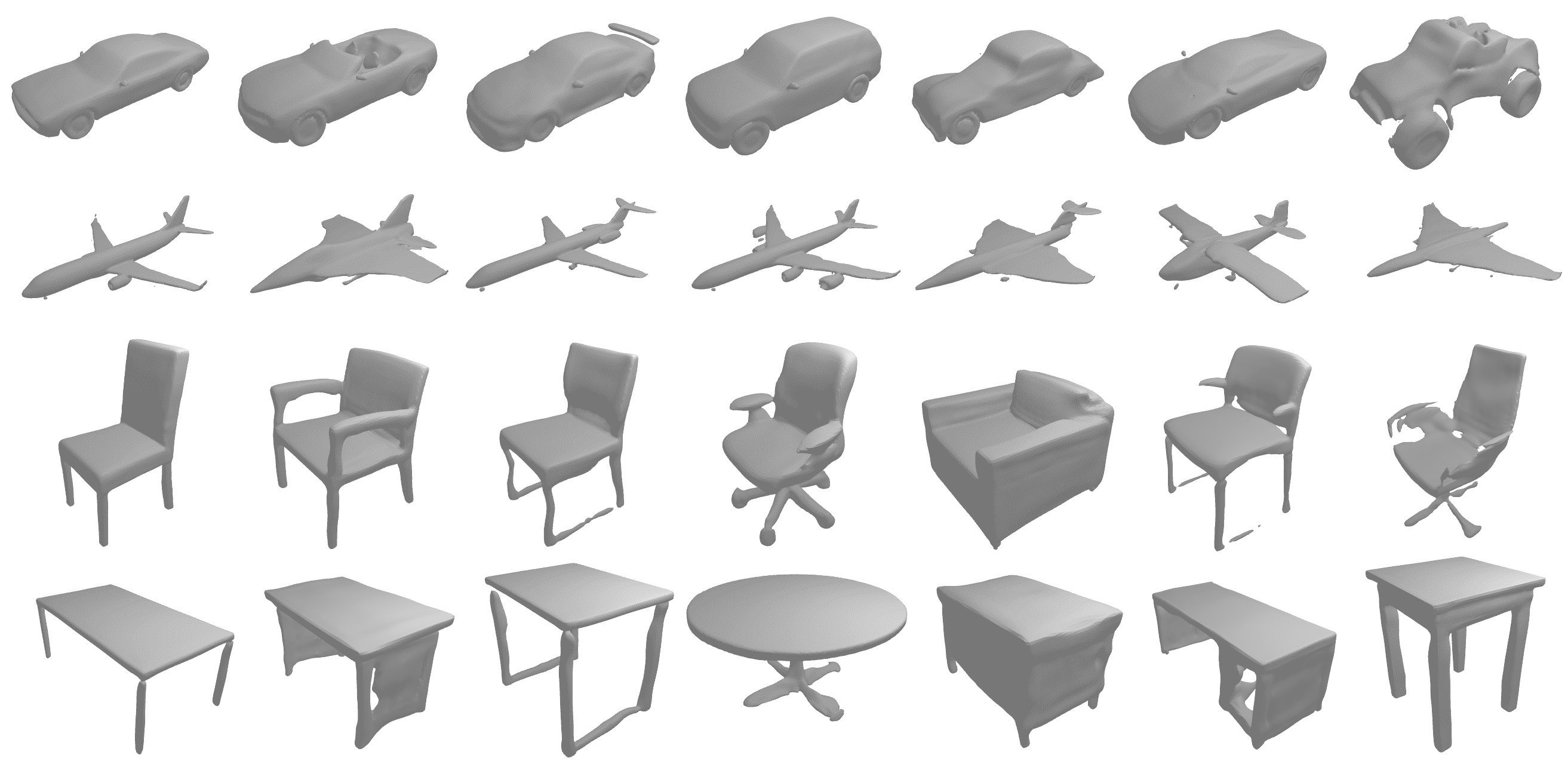}
	\caption{Sampled shapes from our model. 
	\label{fig:sample}}
\end{figure*}

\paravspace
\paragraph{Latent Space Sampling.} We show shape sampling results using the model trained with Eq.~(\ref{eq:all2}) in Fig.~\ref{fig:sample}. Our model can well capture the distribution of 3D shapes and generate new shapes. 

\begin{figure}[t]
	\small
	\centering
	\includegraphics[width=1.0\columnwidth]{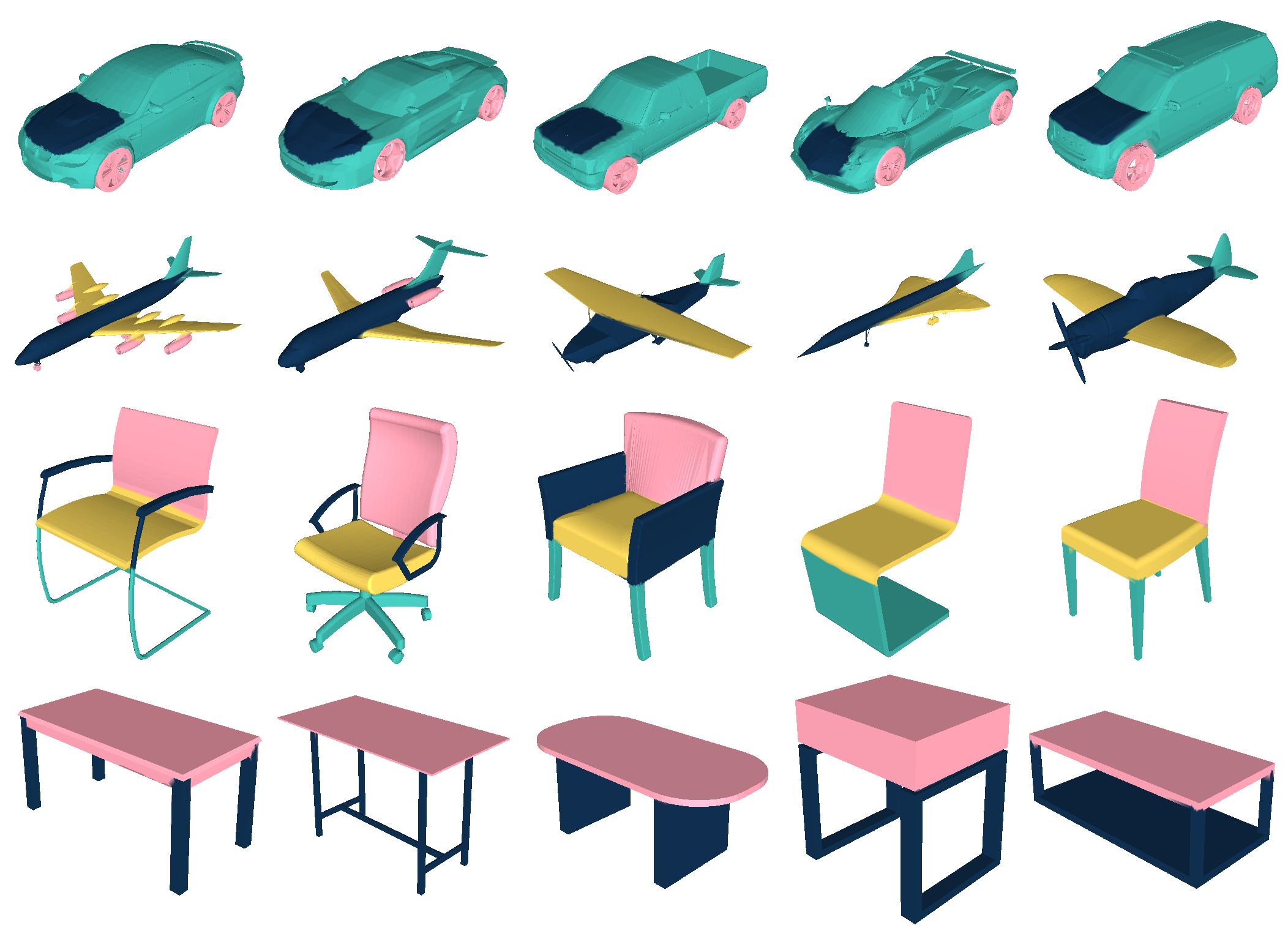}
	\caption{Source shapes of each category used for label transfer evaluation.
	\label{fig:trans_gt}}
\end{figure}

\begin{figure}[t]
	\small
	\centering
	\includegraphics[width=1.0\columnwidth]{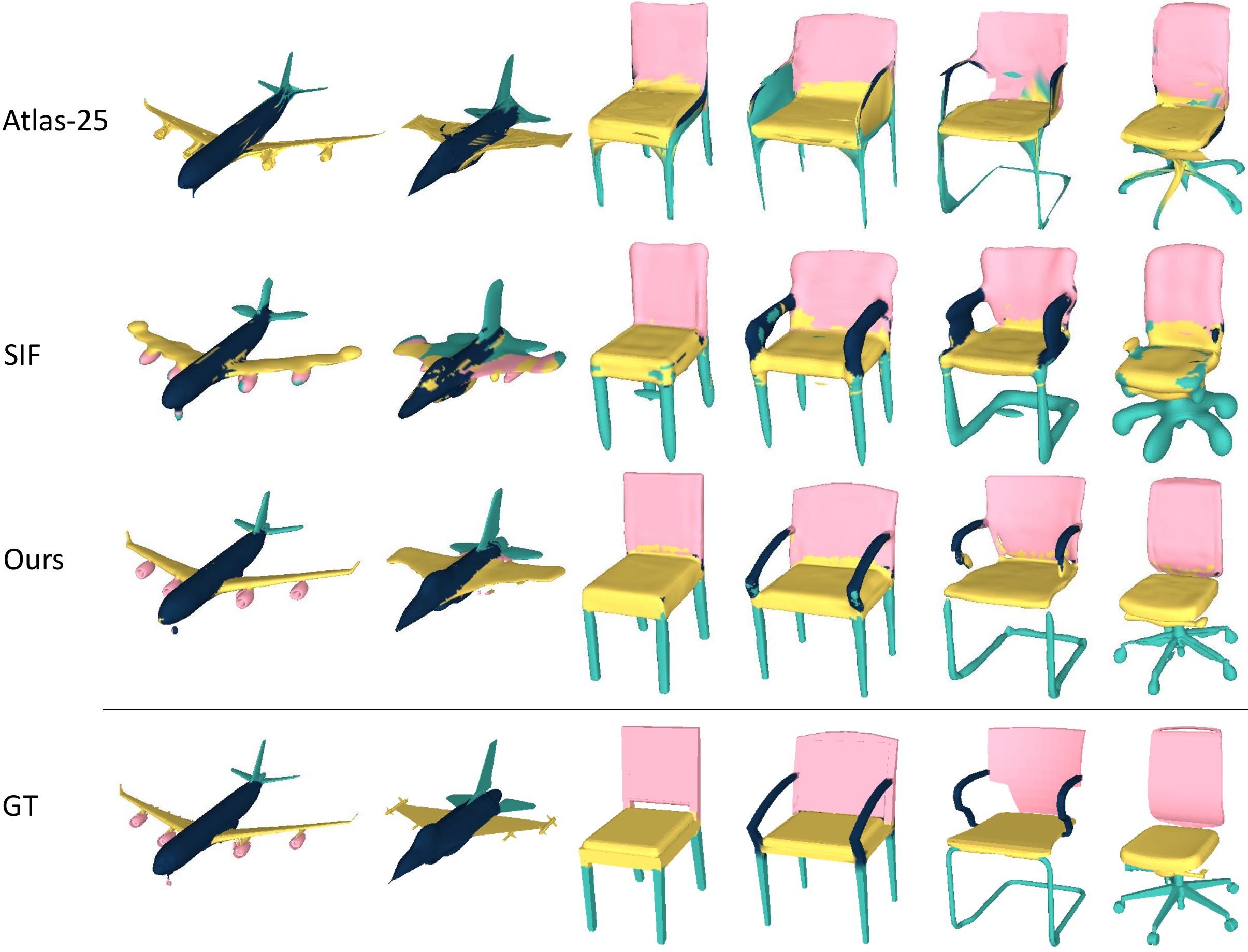}
	\caption{Quatitative comparison on label transfer task.\label{fig:seg_compare}}
	\vspace{-1pt}
\end{figure}

\section{More Details of Label Transfer}
\paragraph{Source Shapes.} Figure~\ref{fig:trans_gt} shows the source shapes used to conduct the label transfer experiment. We manually select 5 sources for each category characterizing various structures in the shape distribution. 

\paravspace
\paragraph{Closest Point.} In the naive closest point method, we label each point by direct label voting using its 10 nearest labeled points in the original shape space.

\paravspace
\paragraph{AtlasNet.} For AtlasNet, we train individual model for each category using two different settings: one using a sphere mesh as the template, dubbed Atlas-sph., and the other using 25 square meshes as the template, dubbed Atlas-25. 

For each trained model, we first fit the template to all source shapes, and label each grid vertex on the template using label voting of its 5 nearest labeled points on 5 source shapes respectively. Then, given unlabeled points on a target shape, we fit the template to that shape, and label all points using label voting of 10 nearest grid vertices on the template.

\paravspace
\paragraph{AtlasNet-v2.} For AtlasNet-v2, we train individual model for each category using the Patch Deformation module with MLP adjustment. For trained models, we follow the same step as for AtlasNet to transfer labels to target shapes.

\paravspace
\paragraph{SIF.} We train SIF for each category using a template of 100 implicit kernels as in \cite{genova2019learning}. To conduct label transfer, we follow \cite{genova2019learning} to first calculate the template coordinates for all point on different shapes. The template coordinates has a dimension of 300 (100 kernels each with a 3-dimensional coordinates). Then, given an unlabeled point on the target shape, we find its 10 nearest labeled points in the template coordinate system and conduct label voting. 

\paravspace
\paragraph{ShapeFlow.} We use the model trained on chair category provided by the authors\footnote{\href{https://github.com/maxjiang93/ShapeFlow}{https://github.com/maxjiang93/ShapeFlow}.}. To conduct label transfer, we deform surface points of source shapes and target shapes into the template space, find 10 nearest labeled source points for each target point, and then conduct label voting.

\paravspace
\paragraph{DualSDF.} We use models trained on plane and chair categories provided by the authors\footnote{\href{https://github.com/zekunhao1995/DualSDF}{https://github.com/zekunhao1995/DualSDF}.}. For the label transfer task, we label each primitive using label voting of its 5 nearest labeled points on 5 source shapes respectively. Then, given an unlabeled point on a target shape, we assign it with the label of its closest primitive.

\paravspace
\paragraph{Qualitative Results.} Figure~\ref{fig:seg_compare} shows the visual results of label transfer task.

\begin{figure}[t]
	\small
	\centering
	\includegraphics[width=1.0\columnwidth]{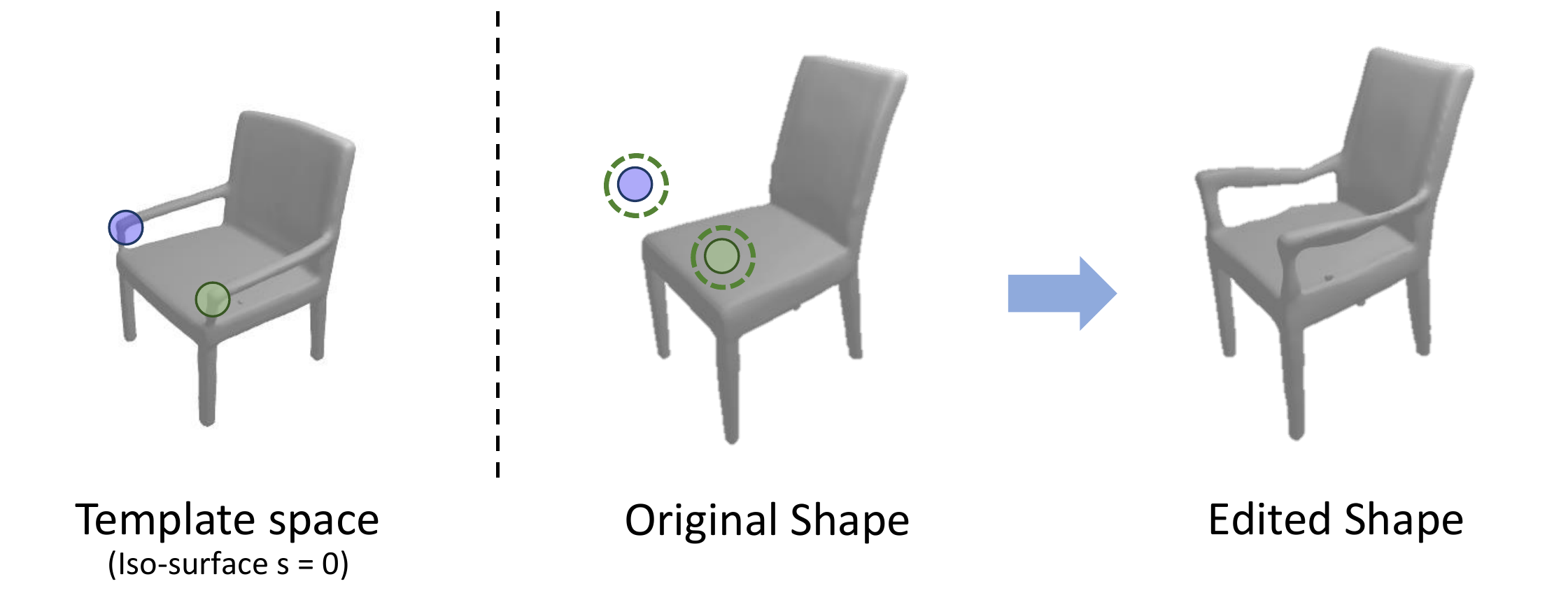}
	\caption{Illustration of adding new structures to a given shape.
	\label{fig:edit_detail}}
	\vspace{-5pt}
\end{figure}

\section{More Details of Shape Editing}

\paragraph{Implementation Details.} We achieve shape editing via Eq.~(\ref{eq:edit}) as described in the main paper. Given a shape with latent code $\alpha$, we initialize the variable $\hat{\alpha}$ to $\alpha$ at the beginning of the optimization. Balancing weight for three terms in Eq.~(\ref{eq:edit}) are set to $1$, $1$, and $5$ respectively. We use Adam optimizer with a learning rate of $1e-4$, and update $\hat{\alpha}$ with 1,000 iterations. The optimization process takes about 10 seconds.

\paravspace
\paragraph{Adding New Structures.} To add new structures to a given shape, we select $p_1'$ defined in Eq.~(\ref{eq:edit}) in the template space, as is shown in Fig.~\ref{fig:edit_detail}. Then, we follow the same optimization process to learn $\hat{\alpha}$ for the new shape as described in the above paragraph. 

\begin{figure*}[t]
	\small
	\centering
	\includegraphics[width=0.97\textwidth]{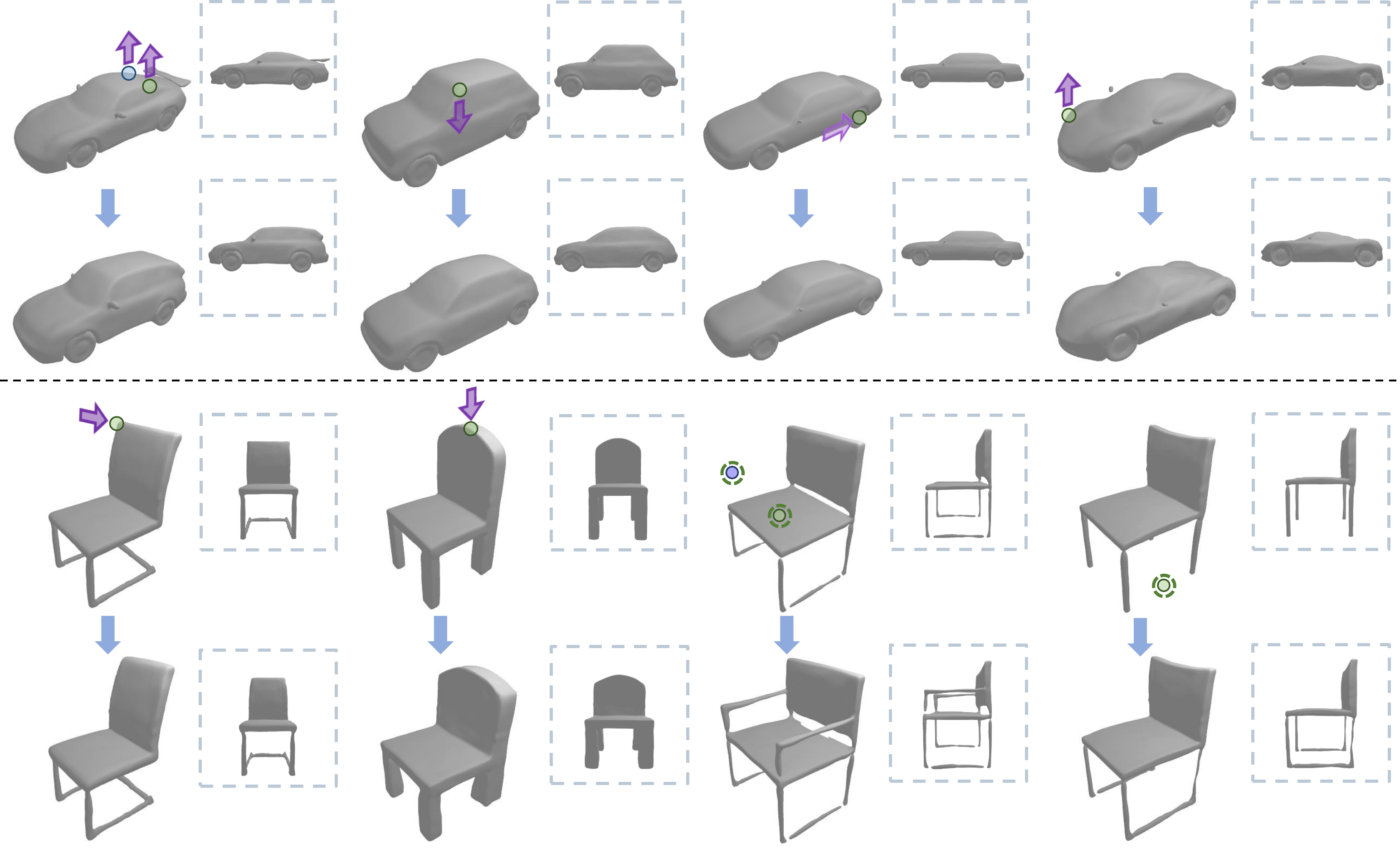}
	\caption{More shape editing results.
	\label{fig:edit_more}}
\end{figure*}

\paravspace
\paragraph{More Editing Results.} We show more shape editing results in Fig.~\ref{fig:edit_more}. Our method is able to move selected points on the shape to desired position or add new structures. Features of the original shapes can be preserved after editing.

\end{document}